\documentclass[letterpaper]{article} 
\usepackage[]{aaai25}  
\nocopyright
\usepackage{times}  
\usepackage{helvet}  
\usepackage{courier}  
\usepackage[hyphens]{url}  
\usepackage{graphicx} 
\usepackage{natbib}  
\usepackage{caption} 
\usepackage[ruled]{algorithm2e} 
\usepackage{algorithmic}

\usepackage{booktabs} 
\usepackage{multirow}
\usepackage{makecell}
\usepackage{cuted}
\usepackage{amsmath}
\usepackage{newfloat}
\usepackage{listings}
\DeclareCaptionStyle{ruled}{labelfont=normalfont,labelsep=colon,strut=off} 
\title{GenesisTex2: Stable, Consistent and High-Quality Text-to-Texture Generation}
\author{
    Jiawei Lu\textsuperscript{\rm 1,2}\footnotemark[1]\footnotemark[2], Yingpeng Zhang\textsuperscript{\rm 2}\footnotemark[2]\footnotemark[3], Zengjun Zhao\textsuperscript{\rm 2}, He Wang\textsuperscript{\rm 3}, Kun Zhou\textsuperscript{\rm 1}, Tianjia Shao\textsuperscript{\rm 1}\footnotemark[3]
}

\affiliations{
    \textsuperscript{\rm 1} State Key Lab of CAD\&CG, Zhejiang University\\
    \textsuperscript{\rm 2} R\&D Efficiency and Capability Department, Tencent IEG\\
    \textsuperscript{\rm 3} University College London
}

\usepackage{bibentry}
\let\oldtwocolumn\twocolumn
\renewcommand\twocolumn[1][]{%
    \oldtwocolumn[{#1}{
    \begin{center}
           \includegraphics[width=0.9\textwidth]{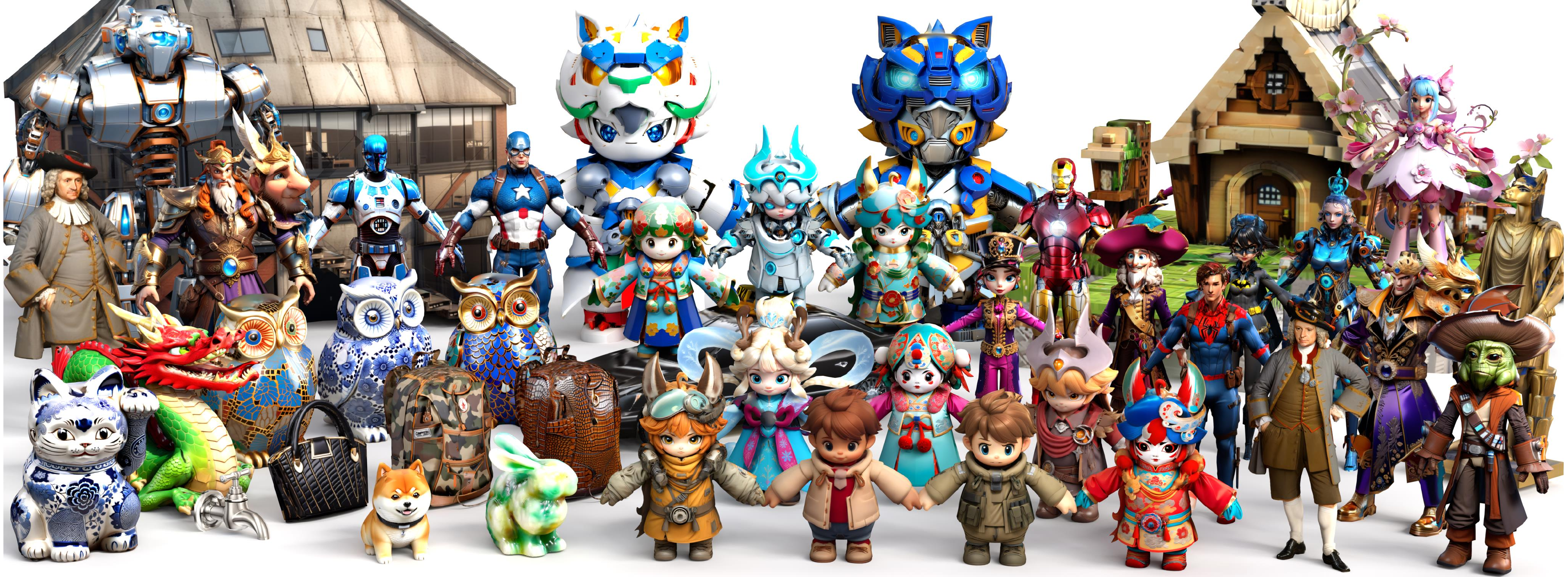}
        \label{fig:teaser}
        \end{center}
    }]
}

\begin{document}
\maketitle

\renewcommand{\thefootnote}{\fnsymbol{footnote}} 
\setcounter{footnote}{0} 
\footnotetext[1]{Work was done during an internship at Tencent IEG.}
\footnotetext[2]{Equal contribution.}
\footnotetext[3]{Corresponding author.}

\begin{abstract}
Large-scale text-guided image diffusion models have shown astonishing results in text-to-image (T2I) generation. However, applying these models to synthesize textures for 3D geometries remains challenging due to the domain gap between 2D images and textures on a 3D surface. Early works that used a projecting-and-inpainting approach managed to preserve generation diversity but often resulted in noticeable artifacts and style inconsistencies. While recent methods have attempted to address these inconsistencies, they often introduce other issues, such as blurring, over-saturation, or over-smoothing. To overcome these challenges, we propose a novel text-to-texture synthesis framework that leverages pretrained diffusion models. We first introduce a local attention reweighing mechanism in the self-attention layers to guide the model in concentrating on spatial-correlated patches across different views, thereby enhancing local details while preserving cross-view consistency. Additionally, we propose a novel latent space merge pipeline, which further ensures consistency across different viewpoints without sacrificing too much diversity. Our method significantly outperforms existing state-of-the-art techniques regarding texture consistency and visual quality, while delivering results much faster than distillation-based methods. Importantly, our framework does not require additional training or fine-tuning, making it highly adaptable to a wide range of models available on public platforms.
\end{abstract}

\section{Introduction}
Digital assets are essential for the gaming, film, and animation industries. The role of textures is pivotal, as they influence the visual effects and aesthetics. However, creating appealing textures takes considerable effort, even for professionals. Recently, diffusion models trained on billions of image-text pairs have enabled users to generate stunning images from text prompts. However, applying this approach to texture synthesis faces significant challenges, primarily due to: 1) a lack of high-quality text-labeled training data for textures and 2) a domain gap between 2D images and 3D surface textures.
Therefore, most methods of text-guided texture generation circumvent the limitations by employing pretrained 2D text-to-image diffusion models. However, creating 3D consistent textures that maintain high quality remains a significant challenge, even with geometric guidance like Depth maps in ControlNet.

Existing approaches typically navigate a trade-off between single-image quality and multi-view consistency, falling into two main categories.
The first group optimizes an underlying 3D structure based on Score Distillation Sampling~\cite{poole2022dreamfusion, lin2023magic3d, wang2024prolificdreamer}. However, these optimization-based methods are often time-consuming and struggle to match the diversity and quality of text-to-image generation.
The second group generates images from various viewpoints to create the final texture in an optimization-free fashion. This can be achieved through sequential inpainting~\cite{chen2023text2tex,richardson2023texture} or a multi-view diffusion approach~\cite{liu2023text, gao2024genesistex}. Our method falls in this category.

We tackle the challenges of achieving both consistency and quality by introducing a cross-view local attention technique and a latent space merge pipeline specifically designed for the text-to-texture task, using only pretrained T2I models. 
For the local attention, we input the 3D mesh and construct dense patch-level weight matrices based on the 3D locations of patches across different views. Patches that are closer in 3D receive higher weights, while farther ones get lower weights. The weight matrices are then incorporated into the self-attention layers during diffusion to amplify or attenuate the effect of specific patches, thereby enhancing local details and improving the consistency of multi-view images.  
Additionally, we design a latent space merge framework to ensure consistent and high-quality texture synthesis.  
Finally, we propose an efficient texture completion algorithm to fill uncolored UV pixels caused by self-occlusion. The algorithm approximates color dilation in surface space by discretizing the UV into sub-UV islands.

Our contributions can be summarized as follows:
\begin{itemize}
\item We propose a novel local attention mechanism for pretrained T2I models, which leverages 3D priors and establishes patch correspondences across different views.
\item We design a framework that incorporates a latent merge pipeline and an efficient texture dilation algorithm in surface space, enabling a stable generation of consistent and high-quality textures.
\item We have conducted extensive evaluations on a variety of 3D objects. The evidence demonstrates that our approach significantly surpasses the performance of the baseline methods by better preserving the generative potential of the original T2I models in aspects of details and color richness while maintaining multi-view consistency. 
\end{itemize}

\section{Related Works}

\begin{figure*}
  \centering
  \includegraphics[width=\linewidth]{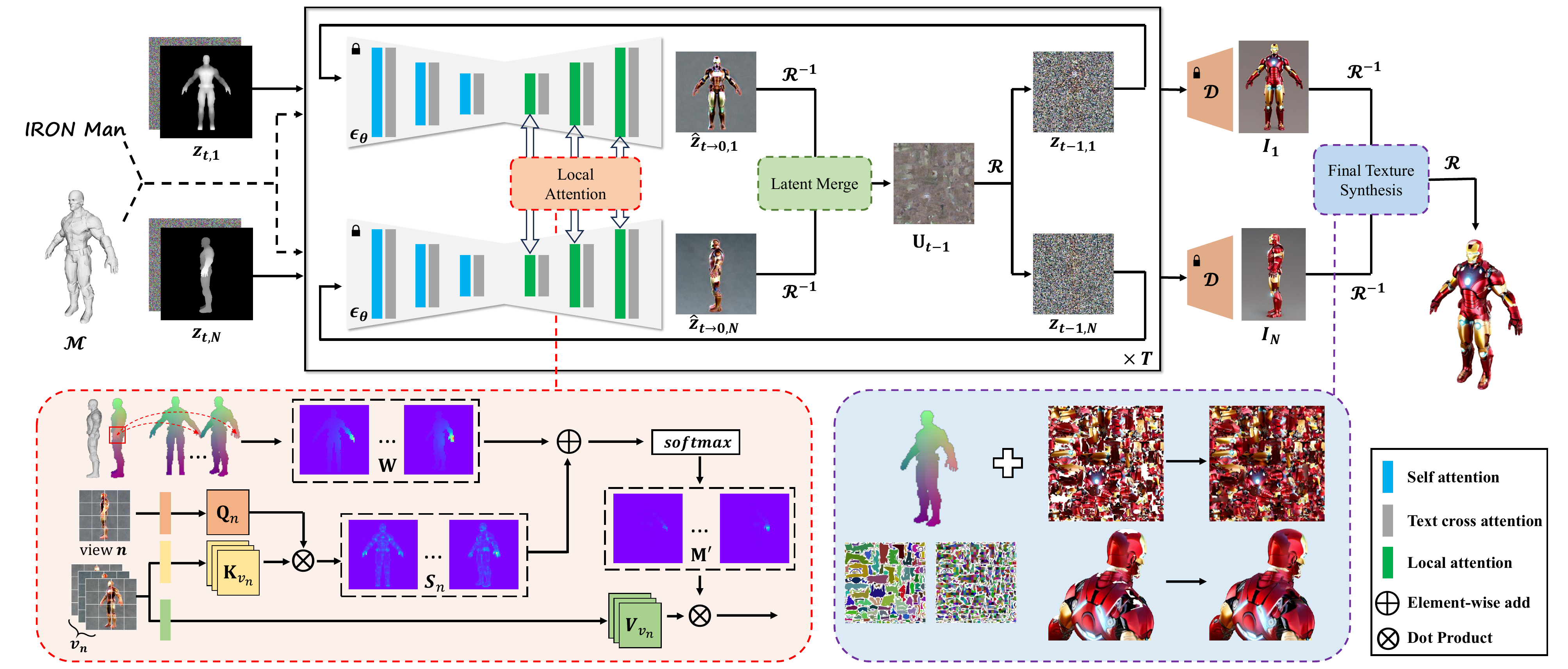}
  \caption{
    Given a mesh and a textual prompt, we aim to produce textures that well depict the prompt and suit the shape. To achieve this, we propose a local attention technique in Sec.~\ref{sec:method-3d aware}, which enhances local details by reweighing the original self-attention layers based on the 3D shape. In addition, we introduce a framework for consistent texture synthesis in Sec.~\ref{sec:method3.3}, which includes a latent merge pipeline and an efficient texture dilation algorithm, enabling the stable generation of consistent and high-quality textures.
  }
  \vspace{-0.15in}
  \label{fig:pipeline}
\end{figure*}

\subsection{Text-to-Image Diffusion Models}
Diffusion models are a class of generative models that use Markov chains to transform random noise into high-quality visuals sequentially. A pioneering work, GLIDE~\cite{nichol2021glide}, is the first to employ diffusion models for generating images in pixel space while supporting text conditioning by adopting classifier-free guidance. Following GLIDE~\cite{nichol2021glide}, Imagen \cite{saharia2022photorealistic} integrates diffusion models for high-resolution text-guided image generation. DALLE-2~\cite{ramesh2022hierarchical} leverages CLIP~\cite{radford2021learning}, a popular model that aligns texts and images to generate images from CLIP latent space. Stable diffusion is a landmark work built upon Latent Diffusion Model (LDM)~\cite{rombach2022high} trained on a large-scale text-image dataset~\cite{schuhmann2022laion}, which proposes to adapt the diffusion process in latent space to further reduce computational cost. Besides text conditioning, various flexible conditions have been introduced for image generation such as ControlNet~\cite{zhang2023adding} and T2I-Adapter~\cite{mou2024t2i}. These control methods aim to generate results that align with a given spatial condition, such as depth or normal images, which can be either predicted from input images or rendered from 3D meshes, supporting mesh-guided image generation.
\subsection{Text-driven 3D Generation}
Many recent studies~\cite{jun2023shap, hong2023lrm, huang2023textfield3d, xu2023dmv3d, nichol2022point} attempt to replicate the success of 2D diffusion models in text-guided 3D content generation, after the supervision of text-paired 3D data. A common constraint of these methods lies in the scarcity of publicly available labeled 3D data. As such, rather than direct leaning a 3D diffusion model, many works resort to using pretrained 2D image diffusion models for 3D tasks~\cite{gao2024genesistex, cao2023texfusion, chen2023text2tex, liu2024one, liu2023zero, long2023wonder3d,shi2023mvdream}.  
Pioneering works~\cite{poole2022dreamfusion, wang2023score} suggest optimizing a 3D representation(E.g., NeRF) by distilling from 2D diffusion models. Subsequent research~\cite{lin2023magic3d,metzer2023latent} further improved such text-to-3D distillation methods in various aspects. A recent remarkable work~\cite{wang2024prolificdreamer} proposed a technique called Variational Score Distillation (VSD) that further enriches the details and diversity. 
Another line of work~\cite{shi2023mvdream, liu2023syncdreamer, tsalicoglou2023textmesh} typically fine-tune a multi-view diffusion model by incorporating camera directions to image diffusion models and simultaneously generate multi-view images. Zero-1-to-3~\cite{liu2023zero} first attempts to leverage 3D data and camera parameters to fine-tune pretrained 2D diffusion models for 3D-consistent novel view synthesis. MVDream~\cite{shi2023mvdream} and SyncDreamer~\cite{liu2023syncdreamer} share a similar idea to improve consistency by fine-tuning attention layers in 2D diffusion models using 2D and 3D data. 

\subsection{Mesh-guided Texture Synthesis}
Beyond generating 3D objects using text prompts, creating textures for given meshes is also a critical and challenging task with various applications. Initial studies~\cite{oechsle2019texture, siddiqui2022texturify, yu2021learning, chen2022auv} have shown promising results using GANs. However, their application is limited to specific categories. In contrast, many recent works on mesh-guided text-to-texture synthesis have achieved broader applicability by leveraging large-scale pretrained diffusion models. These methods typically employ strategies such as sequentially generation and inpainting~\cite{chen2023text2tex, richardson2023texture, cao2023texfusion}, multi-view diffusion~\cite{gao2024genesistex, liu2023text} or score distillation~\cite{chen2023scenetex, metzer2023latent, youwang2023paint}. 

\section{Method}

\label{sec:method-overview}
Given a mesh $\mathcal{M}$ and a textual prompt $\mathcal{P}$, our goal is to produce a texture $\mathcal{T}$ that well depicts the prompt and suits the shape with high quality. An overview of our pipeline is shown in Fig.~\ref{fig:pipeline}. In this section, We first introduce preliminaries on image space diffusion models and define notations for rendering. Next, we provide details on how to adapt the local attention to the diffusion process to improve the local details in the generated images while preserving consistency. Then, we illustrate our latent merge pipeline, which is combined with the local attention mechanism and ensures the consistency. The final texture can be obtained by inverse rendering and merging the generated multi-view images. 

\subsection{Preliminary}
\label{sec:method-Preliminary}
\subsubsection{2D Image Diffusion models}
In this paper, we employ Stable Diffusion~\cite{rombach2022high}. Stable Diffusion is a latent diffusion model that operates in the latent space of an autoencoder $\mathcal{D}(\mathcal{E}(\cdot))$, where $\mathcal{E}$ and $\mathcal{D}$ represent the encoder and decoder, respectively.
For a given image $I$ with its corresponding latent feature $\mathbf{z}_0=\mathcal{E}(I)$, the DDPM forward process\cite{ho2020denoising} iteratively adds gaussian noise to $\mathbf{z}_0$.
\begin{equation}
  q(\mathbf{z}_t|\mathbf{z}_{t-1})=\mathcal{N}(\mathbf{z}_t;\sqrt{\alpha_t}\mathbf{z}_{t-1}, (1-\alpha_t)\mathbf{I}),
\end{equation}
where $t=1,...,T$ is the time step, $q(\mathbf{z}_t|\mathbf{z}_{t-1})$ is the conditional density of $\mathbf{z}_t$ given $\mathbf{z}_{t-1}$, and $\alpha_t$ is hyperparameter. 
In the DDPM backward process, a U-Net $\epsilon_\theta$ is trained to predict the noise and $\mathbf{z}_{t-1}$ can be sampled based on $\mathbf{z}_t$ and prompt $\mathcal{P}$:
\begin{equation}\label{eq:ddpm}
  \mathbf{z}_{t-1}=\frac{\sqrt{\bar\alpha_{t-1}}\beta_t }{1-\bar\alpha_t}\hat{\mathbf{z}}_{t\rightarrow0} + \frac{(1-\bar\alpha_{t-1})(\sqrt{\alpha_t}\mathbf{z}_t+\beta_t\varepsilon_t)}{1-\bar\alpha_t},
\end{equation}
where $\alpha_t$ and $\beta_t=1-\alpha_t$ are pre-defined hyperparamters,  $\hat{\mathbf{z}}_{t\rightarrow0}$ is the denoised estimation at time step $t$, $\epsilon_\theta(\mathbf{z}_t, t, \mathcal{P})$ is the predicted noise for $\mathbf{z}_t$, and $\varepsilon_t\sim \mathcal{N}(0,\mathbf{I})$.
We can sample $\mathbf{z}_0$ by iteratively performing denoising  using Eq.~\ref{eq:ddpm} from the standard Guassian noise $\mathbf{z}_T, \mathbf{z}_T\sim\mathcal{N}(0, \mathbf{I})$ with DDPM sampling, and decode to the final generated image by $\mathcal{D}(\mathbf{z}_0)$.


\subsubsection{Rendering Representation}
In this paper, textures are defined in 2D image space in an injective UV parameterization of $\mathcal{M}$, represented as $UV: p \in \mathcal{M} \mapsto (u,v) \in [0,1] ^ 2$. This parameterization can be automatically constructed using tools like \textit{xatlas}~\cite{xatlas2016}. We focus on synthesizing base color maps and disregard any shading effects. Given a mesh $\mathcal{M}$, a texture map $\mathcal{T}$ and a viewpoint $\mathbf{C}$, we use the rendering function $\mathcal{R}$ to get the rendered image $\mathbf{x} = \mathcal{R}(\mathcal{T}; \mathcal{M}, \mathbf{C})$. 
Conversely, the inverse rendering function $\mathcal{R}^{-1}$ is utilized to reconstruct the texture map from the rendered image: $\mathcal{T}^{\prime} = \mathcal{R}^{-1}(\mathbf{x}; \mathcal{M}, \mathbf{C})$. 
For simplicity, we omit $\mathcal{M}$ and $\mathbf{C}$ for $\mathcal{R}$ and $\mathcal{R}^{-1}$ throughout this paper.

\subsection{Local Attention}

\begin{figure}
  \centering
  \includegraphics[width=\linewidth]{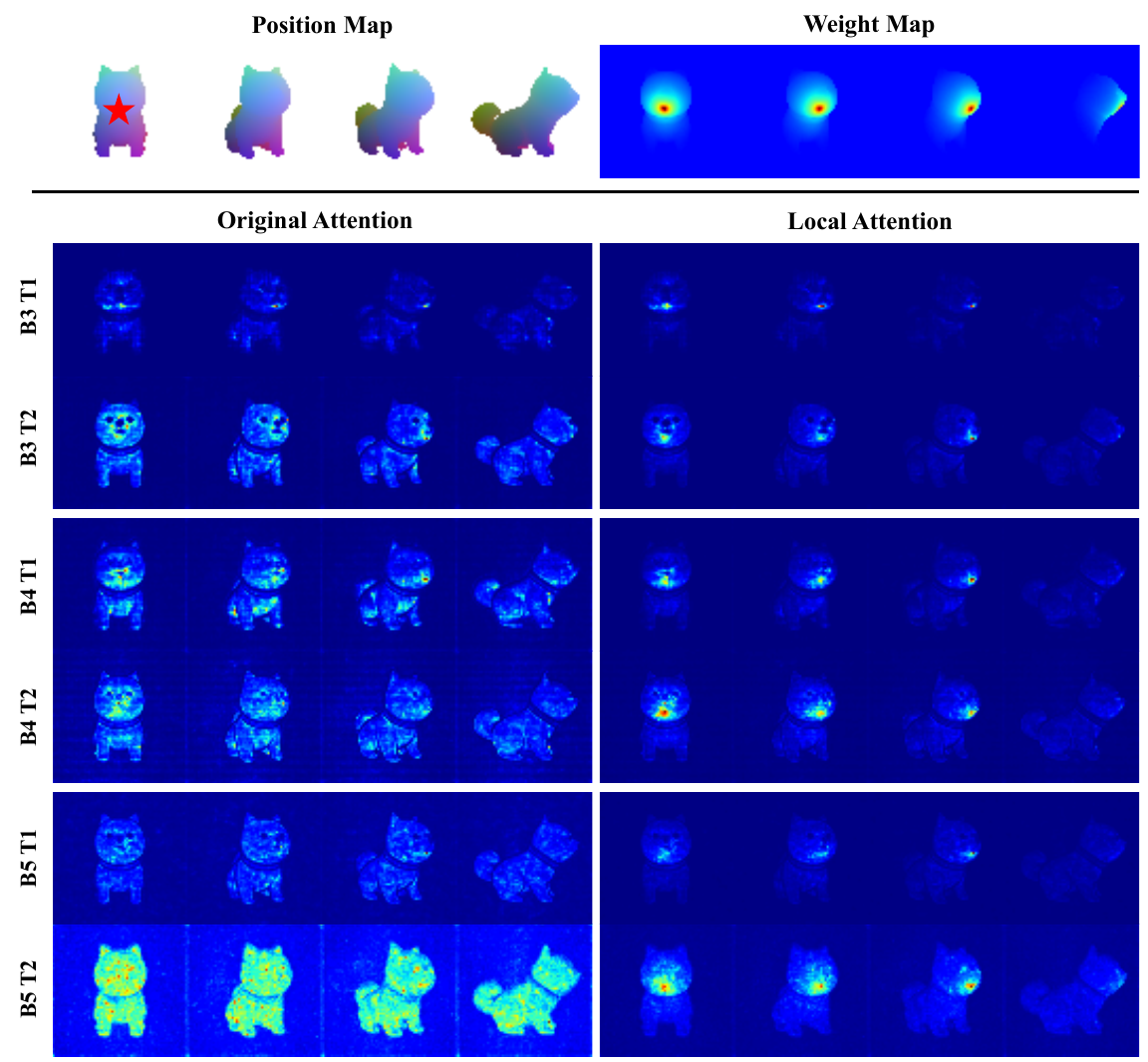}
  \caption{
    A visualization of attention maps concerning the query patch (in red star). The upper part illustrates the rendered position map and calculated weight map. The bottom part shows the attention map of different layers before and after reweighed by the weight map. $B\{i\}T\{j\} $ stands for the $i$-th Block and $j$-th Transformer layer in the output layers. 
  }
  \label{fig:pipeline_attention}
\end{figure}

\label{sec:method-3d aware}
The attention layer is crucial in Stable Diffusion, featuring two types of attention mechanisms: 1) cross-attention, which measures the similarity between the latent features and text embeddings, and 2) self-attention, which can be viewed as patch matching and voting within a single image. In Stable Diffusion, each self-attention layer receives the deep spatial feature $\phi(\mathbf{z}_t)$ of the noisy latent $\mathbf{z}_t$, and linearly projects $\phi(\mathbf{z}_t)$ to the query, key, and value matrices $\mathbf{Q}=l_Q(\phi(\mathbf{z}_t))$, $\mathbf{K}=l_K(\phi(\mathbf{z}_t))$, $\mathbf{V}=l_V(\phi(\mathbf{z}_t))$, where $l_Q, l_K, l_V$ are pretrained linear networks for feature projection. The output of self-attention layers is given by 
$\textit{Softmax}(\frac{\mathbf{Q}\mathbf{K}^T}{\sqrt{d}}) \cdot \mathbf{V}$, where $d$ is a constant representing the dimension of deep features, we omit $\sqrt{d}$ for simplicity in this paper.

Previous works in zero-shot video editing\cite{yang2023rerender, yang2024fresco, khachatryan2023text2video} have demonstrated that modifying the self-attention layers to incorporate cross-frame attention can help regularize style across multiple frames. In texture synthesis, a similar strategy for improving style consistency involves using features from other views as keys and values to perform cross-view attention, as in \cite{gao2024genesistex,liu2023text}. The cross-view attention for view $n$ can be written as: 
\begin{equation}
\textit{cross\_view\_attn}(n) = \textit{Softmax}({\mathbf{Q}_{n}\mathbf{K}_{v_n}^T }) \mathbf{V}_{v_n},
\end{equation}
where $v_n$ is a set of views that attend to the query view $n$. The $\textit{cross\_view\_attn}$ behaves as the original self-attention when $v_n$ contains only $n$.

However, directly adopting this strategy in the diffusion process often leads to a decrease in color diversity and local details in the generated images, as demonstrated in Fig.~\ref{fig:ablation_local_attn}. The root cause of the degradation lies in a reduction of variance in the cross-view attention mechanism, as the predicted feature embedding with the same underlying 3D structure can vary when viewed from different perspectives. This can result in a large attention weight for irrelevant patches, as illustrated in the visualization of attention maps in Fig.~\ref{fig:pipeline_attention}.  In this situation, it becomes necessary to guide the attention module to give greater weight to the same surface area across different viewpoints. This requires considering the correlation of patches among multiple views. Fortunately, we have the input 3D proxy in the texture synthesis task, which naturally builds a strong semantic correspondence between patches of different views. 

Inspired by \cite{hertz2022prompt}, which enables prompt-based image editing by modifying the cross-attention layers in diffusion models, we introduce an attention bias matrix $\mathbf{W}$ to reweigh the original attention produced by the pretrained self-attention layers in Stable Diffusion. Similar to the attention mask mechanism that masks certain words in the cross-attention layers, $\mathbf{W}$, in our case, is used to emphasize or diminish the correlation between specific pairs of query-key patches within the self-attention layers. Unlike the previously mentioned cross-view \textbf{\textit{global}} attention, we refer to our approach as cross-view \textbf{\textit{local}} attention.

We now define the process for calculating the attention bias $\mathbf{W}$. Without loss of generality, let us consider the local attention of the $n$-th query view with attended views denoted as $v_n$. For simplicity, we will omit the subscript $n$ until the end of this section. We render a set of position maps $\{\mathcal{O}\}$ by applying the rendering function $\mathcal{R}(\mathcal{V})$ to each view in $v$, where $\mathcal{V}$ denotes the vertex position of $\mathcal{M}$. Then, we calculate a distance matrix $\mathbf{d}$ based on the rendered position maps $\{\mathcal{O}\}$. Each entry of $\mathbf{d}$ can be calculated using Euclidean distance: $\mathbf{d}_{i,j}=\|\mathcal{O}_i - \mathcal{O}_j\|$ for any location $i \in \{1,...,N_{Q}\}$ and  $j \in \{1,...,N_{K}\}$, where $N_{Q}$ and $N_{K}$ stands for the number of patches in query and key features, respectively. We do not use geodesic distance due to its significant computational cost, particularly for meshes with a large number of vertices. Furthermore, the precision of the distance calculations is inherently limited by the low resolution of the attention maps, making the choice of distance calculation method less critical.

Then, we compute $\mathbf{W}$ by: 
\begin{equation}
\mathbf{W}_{i,j} =
\begin{cases}
    0, & \textit{if}\quad Q_i \in \textit{BG}\, \cap \, K_j \in \textit{BG} \\
    -o\ln(1+r \mathbf{d}_{i,j}), & \textit{if}\quad Q_i \in \textit{FG}\, \cap \, K_j \in \textit{FG} \\
    -\infty, & \textit{else}
\end{cases}
\end{equation}
where $o$ and $r$ are hyper-parameters that determine the distribution of the attention bias, \textit{BG} and \textit{FG} refer to background and foreground patches, respectively. Intuitively, the attention bias approaches $0$ for patch pair located at the same position in 3D and attenuates towards $-\infty$ as the distance increases. We do not reweigh attention between background patches, and to avoid extreme cases, we set a lower bound $\delta$ by applying a clamping operation: $\mathbf{W} = max(\mathbf{W}, \ln(\delta))$. In our experiments, we empirically set $o,r,\delta$ as $2,20$ and $0.1$ to get the best performance.

Given the original similarity $\mathbf{S}=\mathbf{Q}\mathbf{K}_{v}^T$ and attention bias $\mathbf{W}$, we can compute the reweighed attention matrix as follows:
\begin{equation}
\mathbf{M'} = \textit{Softmax}({\mathbf{S}+ \mathbf{W}}),
\end{equation}
where each element $\mathbf{M}'_{i,j}$ is calculated by:
\begin{equation}
 \mathbf{M'}_{i,j} = \frac{e^{\mathbf{W}_{i,j}}e^{\mathbf{S}_{i,j}}}{\sum_j e^{\mathbf{W}_{i,j}}e^{\mathbf{S}_{i,j}}}
\end{equation}

In this way, we manage to manipulate the attention maps by emphasizing on the correspondence of feature patches that are closer in 3D. We empirically find it helpful to replace the original similarity with the weight matrix, to enforce the local appearance consistency, i.e. $\mathbf{M'} = \textit{Softmax}({\mathbf{W}})$. However, the replacement operation can lead to blurring and shape distortion in the late steps. Therefore, we limit the replacement strategy to the early stages of the diffusion process for rough consistency guidance.

\subsection{Consistent Texture Synthesis}
\label{sec:method3.3}
\subsubsection{Latent merge pipeline}
\label{sec:method-latent space merge}
Applying cross-view local attention in the diffusion process can improve the style consistentcy across different views, but it's still insufficient for synthesizing 3D consistent views, i.e., two pixels projected to the same point in 3D have the same value. Directly merging these views will inevitably cause inconsistencies in the final texture, as shown in the first two rows of Fig.~\ref{fig:ablation_latent_merge}. We consider a latent space alignment strategy similar to \cite{liu2023text, gao2024genesistex, kim2024synctweedies} for better cross-view consistency. However, the alignment operation can lead to an over-smoothed appearance and degradation in diversity due to a loss of variation in the alignment process, see Fig.~\ref{fig:comparison}. To overcome these issues while maintaining view consistency, we introduce a novel latent merge pipeline. 

\begin{figure}
  \centering
  \includegraphics[width=\linewidth]{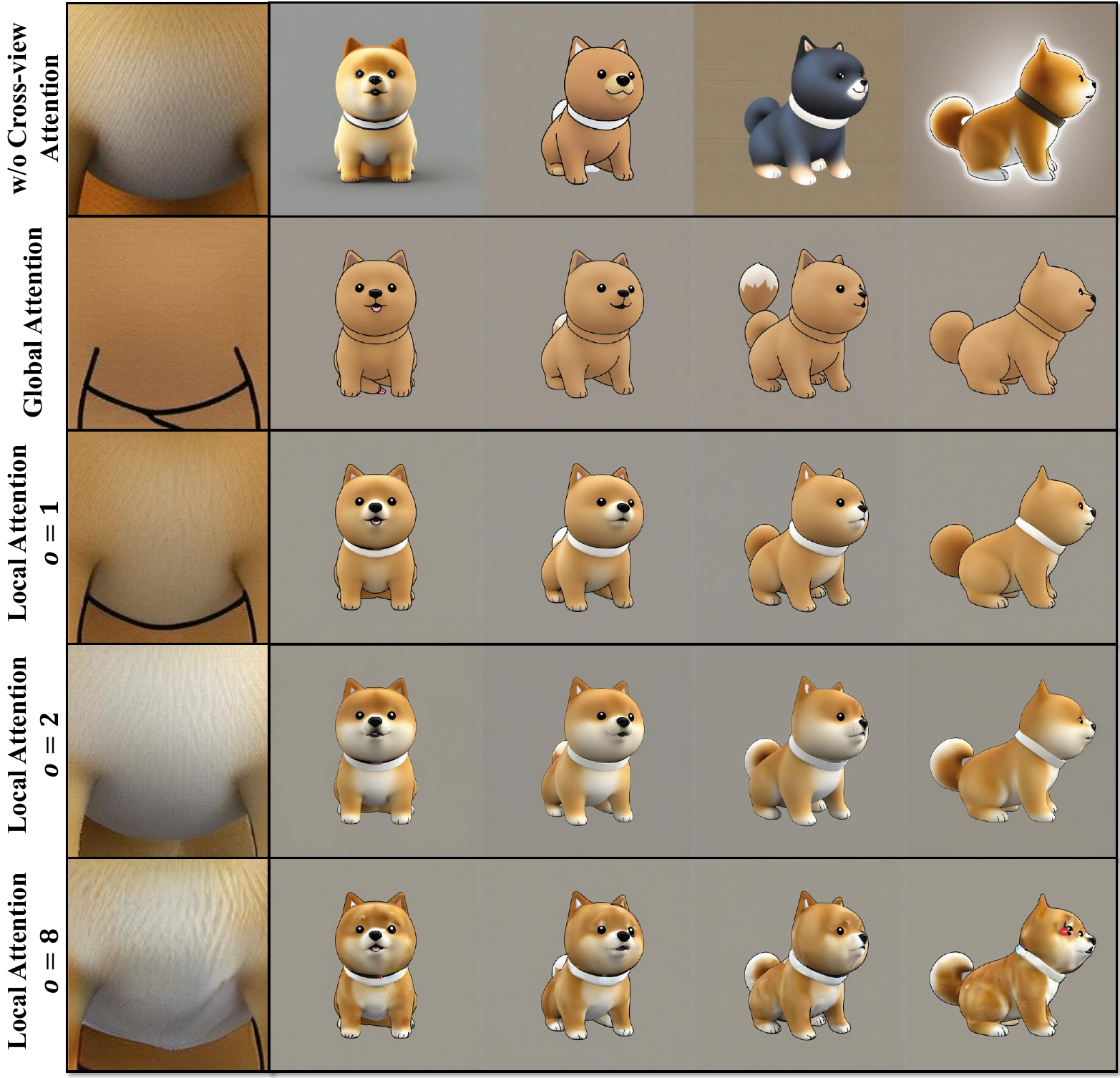}
  \caption{
    Results of different attention mechanisms for 4-view diffusion with prompt: \textit{A cute shiba inu dog}. Images in row 1 are generated without cross-view attention and exhibit no consistency. Results using Global Attention (row 2) are consistent but lose color diversity and details. Images with Local Attention (row 3-5) show improvements in diversity and details, all while maintaining a significant level of cross-view consistency. We find that setting $o=2$ achieves better diversity while eliminating artifacts with $o=8$. 
  }
  \vspace{-0.15in}
  \label{fig:ablation_local_attn}
\end{figure}

Specifically, we first initialize a set of noisy latent for each view by $\{\mathbf{z}_{T,n} \sim\mathcal{N}(0, \mathbf{I})\}_{n=1}^{N}$ and an initial latent texture $\mathbf{U}_T \sim\mathcal{N}(0, \mathbf{I})$ at the beginning of denoising process. At each denoising step $t$, our goal is to predict 3D consistent $\mathbf{z}_{t-1,n}$ from $\mathbf{z}_{t,n}$. We first obtain the denoised prediction $\hat{\mathbf{z}}_{t\rightarrow0, n}$ in image space by:
\begin{equation}
\label{eq:}
    \hat{\mathbf{z}}_{t\rightarrow0,n}=(\mathbf{z}_{t,n}-\sqrt{1-\alpha_{t}}\epsilon_\theta(\mathbf{z}_{t,n}, t, \mathcal{P}, d_n))/\sqrt{\alpha_{t}},
\end{equation}
where $d_n$ is the depth condition for ControlNet at view $n$. 

We then apply inverse rendering to obtain the per-view partial latent textures by: 
\begin{equation}
    \hat{\mathbf{U}}_{t\rightarrow0, n} = \mathcal{R}^{-1}( \hat{\mathbf{z}}_{t\rightarrow0, n}).
\end{equation}

Note that the partial textures do not exhibit 3D consistency at this moment. One way is to aggregate them into a canonical one by averaging. However, trivially averaging the partial textures of different views can lead to a loss of high-frequency details and color diversity. Hence, we propose to merge them in a view-dependent way:
\begin{equation}
    \hat{\mathbf{U}}_{t\rightarrow0} = \frac{\sum_{n=1}^{N}{\omega_{t,n}\mathcal{R}^{-1}(\mathbf{N}_{n}) \odot \hat{\mathbf{U}}_{t\rightarrow0, n}}} {\sum_{n=1}^{N}{\omega_{t,n}\mathcal{R}^{-1}(\mathbf{N}_{n})}},
\end{equation}
where $\mathbf{N}_{n}$ is the cosine similarity map rendered at view point $C_{n}$ with each pixel representing the cosine similarity between the normal vector of the 3D point and the reversed view direction. The term $\omega_{t,n}$ denotes the weight for view $n$ at time step $t$. $\omega_{t,n}$ is set to $1$ at time step $T$ and is then linearly interpolated to $\max(|\cos\theta|^{\gamma}, \omega_{min})$ at time step $t'$, where $\theta$ is the angle between $C_n$ and $C_0$, and $\gamma$ is a hyperparameter that balances the influence of different views. Intuitively, this approach ensures that at the beginning of the diffusion process, different views are merged with similar weights, promoting style consistency. As the diffusion progresses, each texel becomes predominantly influenced by a single view, effectively preserving diversity and preventing the loss of high-frequency details.

After merging the denoised partial textures into a single one, we can update the latent texture $\mathbf{U}_{t-1}$ by adding back the variance with Eq.~\ref{eq:ddpm}:
\begin{equation}
    \mathbf{U}_{t-1} = \frac{\sqrt{\overline{\alpha}_{t-1}}\beta_t}{1-\overline{\alpha}_t}\hat{\mathbf{U}}_{t\rightarrow0} + \frac{(1-\overline{\alpha}_{t-1})}{1-\overline{\alpha}_t}(\sqrt{\alpha_t}\mathbf{U}_{t}+\beta_t\varepsilon_t).
\end{equation}
The image space latent $\mathbf{z}_{t-1,n}$ for next step of $t-1$ can be then obtained by blending the rendered foreground latent $\mathcal{R}(U_{t-1}, C_k)$ with the image space latent  $\mathbf{\hat{z}}_{t-1,n}$:
\begin{equation}\label{eq:mapping}
\begin{aligned}
  \mathbf{z}_{t-1, n} &= \mathbf{M}_n \odot \mathcal{R}(\mathbf{U}_{t-1}; C_n) + (1-\mathbf{M}_n) \odot \mathbf{\hat{z}}_{t-1, n},
\end{aligned}
\end{equation}
where $\mathbf{\hat{z}}_{t-1, n}$ can be derived by Eq.~\ref{eq:ddpm}, and $\mathbf{M}_n$ represents the binary foreground mask for viewpoint $C_n$. 

The final denoised $\mathbf{z}_{0,n}$ of each view can be obtained by iterating the denoising steps. We do not perform latent merge in the last 5 steps to prevent artifacts caused by the reprojection of low-resolution latents. 

\subsubsection{Final Texture Synthesis}
\label{sec:method-Final Texture}

To reconstruct the texture map, we first decode the latent of each viewpoint to generate multi-view images $\mathcal{I}_n$ by $\mathcal{D}(\mathbf{z}_{0,n})$. Subsequently, we finalize the texture by:
\begin{equation}
\mathcal{T}_{merge} = \frac{\sum_{n=1}^{N} \omega_n\mathcal{R}^{-1}(\mathbf{N_n}) \odot \mathcal{R}^{-1}(\mathcal{I}_n)}{\sum_{n=1}^{N} \omega_n\mathcal{R}^{-1}(\mathbf{N_n})}
\end{equation}
where $\mathbf{N_n}$ is the similarity mask at viewpoint $C_{n}$ and $\omega_n=\max(|\cos\theta|^{\gamma}, \omega_{min})$.

After merging, the texture map still contains invalid pixels that fail to receive color from any perspective due to self-occlusion. A straightforward approach to address this issue is to expand the valid pixels on the texture map using a flood-fill technique within the image space. However, this naive flood-fill method may propagate colors from pixels that are not adjacent in the 3D space, leading to inaccuracies in the final texture map. An optimal solution involves using geodesic distance, but the computational cost is prohibitively high. Therefore, we introduce a fast texture completion method that approximates color propagation in surface space. The detailed algorithm could be found in the Appendix.

\section{Experiments}
\label{sec:exp}

\begin{figure}
  \centering
  \includegraphics[width=\linewidth]{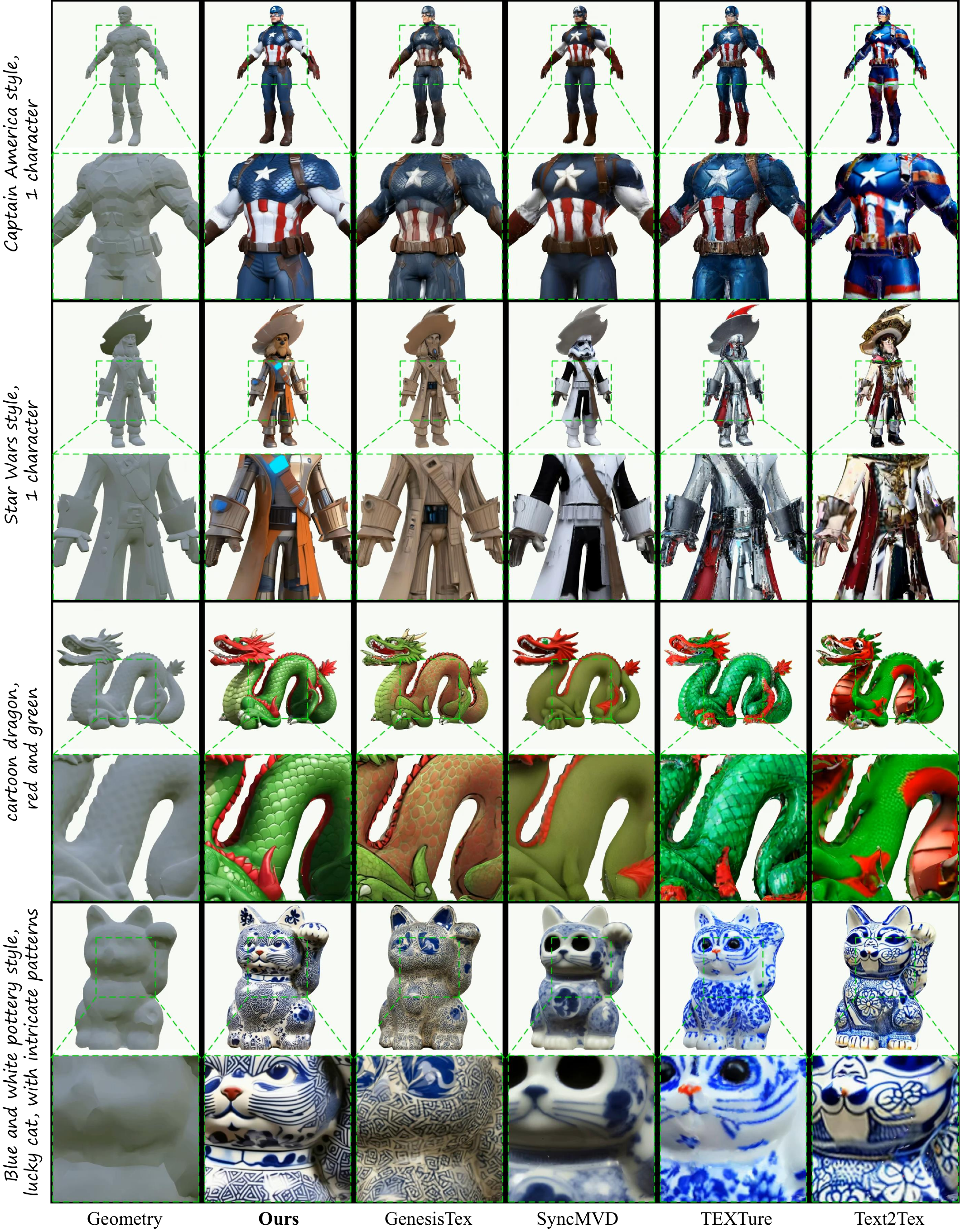}
  \caption{
    Qualitative comparison with different baselines. 
  }
  \label{fig:comparison}
\end{figure}

\subsection{Implementation details}
We test our method on an NVIDIA A800 GPU, and the entire process was able to finish within 1 minute. The diffusion process takes around 50s with 25 denoising steps at a resolution of 1280, and the final texture synthesis stage takes around 2s. The CFG scale is set to 12. We linearly interpolate the view-dependent weight $\omega$ for the first 8 steps. The paramters $\gamma$ and $\omega_{min}$ are set as $8$ and $1e-3$. We adopt SDXL~\cite{podell2023sdxl} as our base model and ControlNet-Depth~\cite{zhang2023adding} trained for SDXL for spatial control. We replace the self-attention layers in the output layers of SDXL by our proposed 3D-aware local attention mechanism in all experiments. 

\subsubsection{Dataset}
The dataset used in evaluation contains 35 meshes with 63 mesh-prompt pairs. The meshes are collected from the publicly open dataset including objarverse~\cite{deitke2023objaverse}, shapenet~\cite{chang2015shapenet}, and stanford 3D Scanning Repository~\cite{turk1994zippered}. We use \textit{Xatlas}~\cite{xatlas2016} to automatically unwrap the UV for all meshes. We normalize all meshes to the range of $[-0.5, 0.5]$ and position the camera at a distance of $2$ meters with the field of view set to $35$ degrees.
To balance time-cost and view coverage, we typically employ $N=8$ fixed viewpoints at angles of $[0,45,90,135,180,225,270,315]$ degrees, evenly distributed around the object of interest.

\begin{table}
\small
\centering

\begin{tabular}{l c c c c c c} 
\toprule
\multirow{2}{*}{Method} & \multirow{2}{*}{\makecell{PS $\uparrow$\\$(\%)$}} &\multirow{2}{*}{\makecell{FID$\downarrow$ }} &\multirow{2}{*}{\makecell{KID$\downarrow$\\$\times 10^{-3}$ }} & \multicolumn{3}{c}{User study (\%)} \\
 & & & &  \makecell[c]{D $\uparrow$} &  \makecell[c]{C  $\uparrow$} & \makecell[c]{Q $\uparrow$}  \\
\midrule
Text2Tex   & 9.7 & 88.1 & 14.2 & 14.3 & 4.5 & 7.0 \\
TEXTure   & 10.2 & 92.2 & 17.1 & 13.3 & 5.7 & 6.7 \\
GenesisTex   & 17.1 & 77.0 & 9.5 & 10.8 & 12.7 & 11.4 \\
SyncMVD    & 13.6  & 85.7 & 10.2 & 3.5 & 5.7 & 8.9 \\
$\textbf{Ours}$ & $\textbf{49.4}$ & $\textbf{66.4}$ & $\textbf{7.3}$ & $\textbf{58.1}$  & $\textbf{71.4}$ & $\textbf{66.0}$ \\
\bottomrule
\end{tabular}

\caption{ \small Quantitative comparisons with baseline methods. Pick Score (PS), Diversity (D), Consistency (C), and Quality (Q).}
\label{tab: comp}
\end{table}



\subsection{Comparisons}
We conduct comparison with four available methods on text-to-texture synthesis, including Text2Tex~\cite{chen2023text2tex}, TEXTure~\cite{richardson2023texture}, SyncMVD~\cite{liu2023text}, GenesisTex~\cite{gao2024genesistex}. We have also compared our method with Meshy-3~\cite{meshy}, a state-of-the-art commercial software that supports generating textures for 3D models using text prompts. The comparison results with Meshy-3 are placed in the Appendix. We strongly recommend readers check the appendix for more details. 
\subsubsection{Qualitative comparisons.}
We compare qualitatively with different baselines in Fig.~\ref{fig:comparison}. GenesisTex~\cite{gao2024genesistex} produces visually reasonable renderings, but they tend to generate less diverse images. TEXTure~\cite{richardson2023texture} and Text2Tex~\cite{chen2023text2tex} lacks multi-view consistency since it operates on each view independently. SyncMVD~\cite{liu2023text} yields visually consistent renderings. However, they tend to get blurry results, see the dragon and lucky cat in Fig.~\ref{fig:comparison}, since the latent averaging operation in their approach leads to a loss of high-frequency details and color diversity.


\begin{figure}
  \centering
  \includegraphics[width=\linewidth]{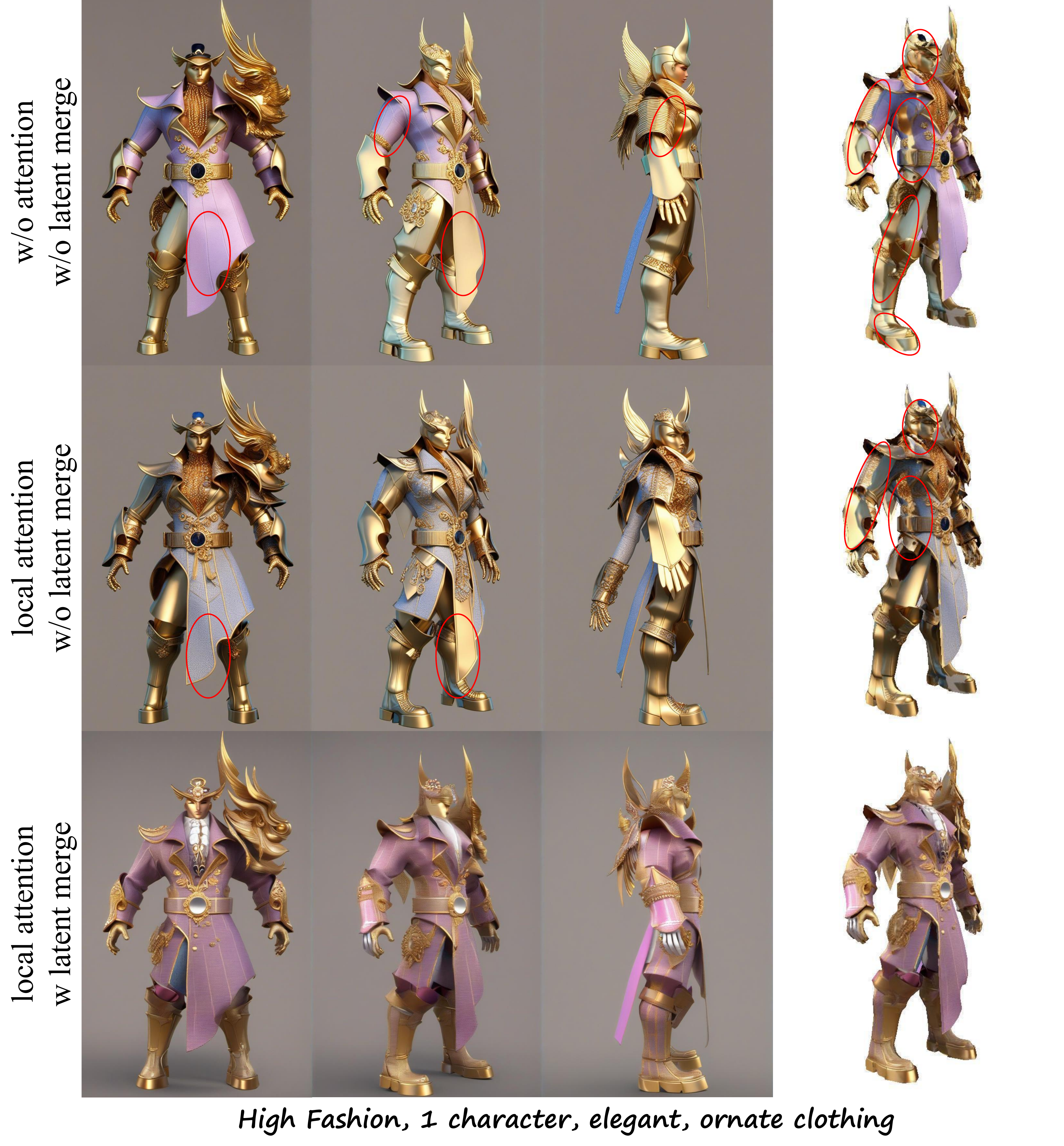}
  \caption{
    Ablation results on local attention and latent merge. 
    The left three columns show the generated images, and the last column depicts the rendered result with synthesized texture.
  }
  \vspace{-0.15in}
  \label{fig:ablation_latent_merge}
\end{figure}
\subsubsection{Quantitative comparisons.}
Following GenesisTex~\cite{gao2024genesistex} and TexFusion ~\cite{cao2023texfusion}, we report FID~\cite{heusel2017gans} and KID~\cite{binkowski2018demystifying} scores. We generate depth maps as conditional images for all meshes by rendering them from 12 different viewpoints, each separated by 30-degree intervals. Using these depth maps and our textual prompts, we sample from pretrained image diffusion model to create a set of ground truth images. Additionally, we render meshes with textures generated by different methods using the same views to get the candidate set. We primary focus on the foreground, and we set the background pixels of all images to white. 

In addition, we also employ \textit{Pick Score}~\cite{kirstain2024pick} to evaluate the visual quality of our texture synthesis results. \textit{Pick Score} is an CLIP-based scoring function trained on large-scale user preference regarding generated images paired with text prompts. For each mesh, we compute the average Pick Score using the same 12-view rendered images employed for calculating the FID, identifying the method with the highest score as the winning approach for that mesh and calculating the winning rate for each method. 

We also conducted a user study to analyze the results across three aspects: 1) consistency, 2) diversity, and 3) overall quality. We render the results of different methods into videos that showcase the textured object from a $360^{\circ}$ rotating view. We randomly pick 15 meshes for each questionnaire. and ask the participants to judge which method matches best for each aspect. Finally, We collected 30 valid answers from professional artists and non-professionals. The whole quantitative results can be found in Tab~\ref{tab: comp}. 
Our method achieves the highest pick score compared to other methods and is preferred by most human evaluators in terms of consistency, diversity, and overall quality.

\subsection{Ablation Studies}
\subsubsection{Effectiveness of local attention}
To investigate the impact of the cross-view local attention, we visualize the decoded multi-view images of different attention strategy in Fig.~\ref{fig:ablation_local_attn} and Fig.~\ref{fig:ablation_latent_merge}. Fig.~\ref{fig:ablation_local_attn} illustrates an example with the prompt \textit{A cute shiba inu dog}. We can discover that the color and pattern of the dog varies a lot across different viewpoints without any cross-view constrain. With global attention, the query view attends to all views in the attention layer and brings higher consistency, but at a cost of losing image details and variance. Our proposed geometry-aware local attention amplifies the local attentions on pixels that are closer in 3D, which not only leads to vivid color and fine-grained details, but also preserves cross-view consistency. Similar in Fig.~\ref{fig:ablation_latent_merge}, the cross-view images are more consistent with local attention than the baseline without cross-view attention. 
\subsubsection{Effectiveness of latent merge pipeline}
We ablate the latent merge pipeline to evaluate the effectiveness of our latent merge strategy in generating consistent textures. As shown in the last column of Fig.~\ref{fig:ablation_latent_merge}, the full pipeline with latent merge exhibits the best consistency compared with baselines in the final renderings. Note how the full method achieves the best multi-view consistency and generates rich details, while the baselines without latent merge exhibit severe inconsistencies. 

\subsection{More Applications}
Our method is designed to be fully compatible with existing Stable Diffusion models without the need for additional training. This makes it readily applicable to a wide range of models available on platforms such as Civitai~\cite{civitai} and HuggingFace~\cite{Huggingface}. Furthermore, our pipeline can be seamlessly integrated with auxiliary models tailored for Stable Diffusion, thereby enriching its versatility in practical scenarios. For instance, we can incorporate the IP-Adapter into our framework to facilitate image-guided texture generation, and leverage various LoRAs to achieve distinct artistic styles. The texturing results with LoRAs and IP-Adapters can be found in the supplementary materials.


\section{Discussions}

\subsubsection{Failure Cases}
Our algorithm employs texture dilation to fill the fully-occluded regions, which may wrongly produce overly smoothed results on these fully-occluded areas which should have complex textures. 
Additionally, the Janus effect is a challenge inherent to methods that utilize pretrained 2D image diffusion models. While this issue is alleviated through the proposed local attention and perspective prompts (as seen in DreamFusion), the inherent bias presented in 2D image diffusion models can still result in unwanted anatomical features. 
\subsubsection{Limitation}
As a common limitation in the field of texture synthesis using pretrained 2D diffusion models, the alignment between the mesh and texture is not perfect, which is largely due to the limited control capabilities of the currently available ControlNets. It could be improved along with the development of more powerful control models.
The baked-in lighting effect is another common limitation in this field, and we will leave it as our future work.

\section{Conclusions}

In this article, we propose a pipeline aiming at generating consistent and high-quality textures for 3D meshes using textual prompts. Our method leverages pretrained Stable Diffusion models without any further training or fine-tuning. This makes it highly versatile, capable of handling a wide range of geometry and texture types, and easily adaptable to various models on model-sharing platforms. We believe this work will advance AI-based texturing and opening up new possibilities for 3D content generation.

\bibliography{main}

\begin{thebibliography}{54}
\providecommand{\natexlab}[1]{#1}

\bibitem[{Bi{\'n}kowski et~al.(2018)Bi{\'n}kowski, Sutherland, Arbel, and Gretton}]{binkowski2018demystifying}
Bi{\'n}kowski, M.; Sutherland, D.~J.; Arbel, M.; and Gretton, A. 2018.
\newblock Demystifying mmd gans.
\newblock \emph{arXiv preprint arXiv:1801.01401}.

\bibitem[{Cao et~al.(2023)Cao, Kreis, Fidler, Sharp, and Yin}]{cao2023texfusion}
Cao, T.; Kreis, K.; Fidler, S.; Sharp, N.; and Yin, K. 2023.
\newblock TexFusion: Synthesizing 3D Textures with Text-Guided Image Diffusion Models.
\newblock In \emph{Proceedings of the IEEE/CVF International Conference on Computer Vision}, 4169--4181.

\bibitem[{Chang et~al.(2015)Chang, Funkhouser, Guibas, Hanrahan, Huang, Li, Savarese, Savva, Song, Su et~al.}]{chang2015shapenet}
Chang, A.~X.; Funkhouser, T.; Guibas, L.; Hanrahan, P.; Huang, Q.; Li, Z.; Savarese, S.; Savva, M.; Song, S.; Su, H.; et~al. 2015.
\newblock Shapenet: An information-rich 3d model repository.
\newblock \emph{arXiv preprint arXiv:1512.03012}.

\bibitem[{Chen et~al.(2023{\natexlab{a}})Chen, Li, Lee, Tulyakov, and Nie{\ss}ner}]{chen2023scenetex}
Chen, D.~Z.; Li, H.; Lee, H.-Y.; Tulyakov, S.; and Nie{\ss}ner, M. 2023{\natexlab{a}}.
\newblock Scenetex: High-quality texture synthesis for indoor scenes via diffusion priors.
\newblock \emph{arXiv preprint arXiv:2311.17261}.

\bibitem[{Chen et~al.(2023{\natexlab{b}})Chen, Siddiqui, Lee, Tulyakov, and Nie{\ss}ner}]{chen2023text2tex}
Chen, D.~Z.; Siddiqui, Y.; Lee, H.-Y.; Tulyakov, S.; and Nie{\ss}ner, M. 2023{\natexlab{b}}.
\newblock Text2Tex: Text-driven Texture Synthesis via Diffusion Models.
\newblock \emph{arXiv preprint arXiv:2303.11396}.

\bibitem[{Chen, Yin, and Fidler(2022)}]{chen2022auv}
Chen, Z.; Yin, K.; and Fidler, S. 2022.
\newblock Auv-net: Learning aligned uv maps for texture transfer and synthesis.
\newblock In \emph{Proceedings of the IEEE/CVF Conference on Computer Vision and Pattern Recognition}, 1465--1474.

\bibitem[{civitai(2024)}]{civitai}
civitai. 2024.
\newblock civitai | The Home of Open-Source Generative AI.
\newblock \url{https://civitai.com/}.

\bibitem[{ComfyUI(2024)}]{ComfyUI}
ComfyUI. 2024.
\newblock ComfyUI.
\newblock \url{https://github.com/comfyanonymous/ComfyUI/}.

\bibitem[{Deitke et~al.(2023)Deitke, Schwenk, Salvador, Weihs, Michel, VanderBilt, Schmidt, Ehsani, Kembhavi, and Farhadi}]{deitke2023objaverse}
Deitke, M.; Schwenk, D.; Salvador, J.; Weihs, L.; Michel, O.; VanderBilt, E.; Schmidt, L.; Ehsani, K.; Kembhavi, A.; and Farhadi, A. 2023.
\newblock Objaverse: A universe of annotated 3d objects.
\newblock In \emph{Proceedings of the IEEE/CVF Conference on Computer Vision and Pattern Recognition}, 13142--13153.

\bibitem[{Gao et~al.(2024)Gao, Jiang, Li, Zhang, and Yu}]{gao2024genesistex}
Gao, C.; Jiang, B.; Li, X.; Zhang, Y.; and Yu, Q. 2024.
\newblock GenesisTex: Adapting Image Denoising Diffusion to Texture Space.
\newblock \emph{arXiv preprint arXiv:2403.17782}.

\bibitem[{Hertz et~al.(2022)Hertz, Mokady, Tenenbaum, Aberman, Pritch, and Cohen-Or}]{hertz2022prompt}
Hertz, A.; Mokady, R.; Tenenbaum, J.; Aberman, K.; Pritch, Y.; and Cohen-Or, D. 2022.
\newblock Prompt-to-prompt image editing with cross attention control.
\newblock \emph{arXiv preprint arXiv:2208.01626}.

\bibitem[{Heusel et~al.(2017)Heusel, Ramsauer, Unterthiner, Nessler, and Hochreiter}]{heusel2017gans}
Heusel, M.; Ramsauer, H.; Unterthiner, T.; Nessler, B.; and Hochreiter, S. 2017.
\newblock Gans trained by a two time-scale update rule converge to a local nash equilibrium.
\newblock \emph{Advances in neural information processing systems}, 30.

\bibitem[{Ho, Jain, and Abbeel(2020)}]{ho2020denoising}
Ho, J.; Jain, A.; and Abbeel, P. 2020.
\newblock Denoising diffusion probabilistic models.
\newblock \emph{Advances in neural information processing systems}, 33: 6840--6851.

\bibitem[{Hong et~al.(2023)Hong, Zhang, Gu, Bi, Zhou, Liu, Liu, Sunkavalli, Bui, and Tan}]{hong2023lrm}
Hong, Y.; Zhang, K.; Gu, J.; Bi, S.; Zhou, Y.; Liu, D.; Liu, F.; Sunkavalli, K.; Bui, T.; and Tan, H. 2023.
\newblock Lrm: Large reconstruction model for single image to 3d.
\newblock \emph{arXiv preprint arXiv:2311.04400}.

\bibitem[{Huang et~al.(2023)Huang, Zeng, Dong, Xu, Xu, Lau, and Zuo}]{huang2023textfield3d}
Huang, T.; Zeng, Y.; Dong, B.; Xu, H.; Xu, S.; Lau, R.~W.; and Zuo, W. 2023.
\newblock Textfield3d: Towards enhancing open-vocabulary 3d generation with noisy text fields.
\newblock \emph{arXiv preprint arXiv:2309.17175}.

\bibitem[{Huggingface(2024)}]{Huggingface}
Huggingface. 2024.
\newblock Huggingface.
\newblock \url{https://huggingface.co/}.

\bibitem[{Jun and Nichol(2023)}]{jun2023shap}
Jun, H.; and Nichol, A. 2023.
\newblock Shap-e: Generating conditional 3d implicit functions.
\newblock \emph{arXiv preprint arXiv:2305.02463}.

\bibitem[{Khachatryan et~al.(2023)Khachatryan, Movsisyan, Tadevosyan, Henschel, Wang, Navasardyan, and Shi}]{khachatryan2023text2video}
Khachatryan, L.; Movsisyan, A.; Tadevosyan, V.; Henschel, R.; Wang, Z.; Navasardyan, S.; and Shi, H. 2023.
\newblock Text2video-zero: Text-to-image diffusion models are zero-shot video generators.
\newblock In \emph{Proceedings of the IEEE/CVF International Conference on Computer Vision}, 15954--15964.

\bibitem[{Kim et~al.(2024)Kim, Koo, Yeo, and Sung}]{kim2024synctweedies}
Kim, J.; Koo, J.; Yeo, K.; and Sung, M. 2024.
\newblock SyncTweedies: A General Generative Framework Based on Synchronized Diffusions.
\newblock \emph{arXiv preprint arXiv:2403.14370}.

\bibitem[{Kirstain et~al.(2024)Kirstain, Polyak, Singer, Matiana, Penna, and Levy}]{kirstain2024pick}
Kirstain, Y.; Polyak, A.; Singer, U.; Matiana, S.; Penna, J.; and Levy, O. 2024.
\newblock Pick-a-pic: An open dataset of user preferences for text-to-image generation.
\newblock \emph{Advances in Neural Information Processing Systems}, 36.

\bibitem[{Laine et~al.(2020)Laine, Hellsten, Karras, Seol, Lehtinen, and Aila}]{Laine2020diffrast}
Laine, S.; Hellsten, J.; Karras, T.; Seol, Y.; Lehtinen, J.; and Aila, T. 2020.
\newblock Modular Primitives for High-Performance Differentiable Rendering.
\newblock \emph{ACM Transactions on Graphics}, 39(6).

\bibitem[{Lin et~al.(2023)Lin, Gao, Tang, Takikawa, Zeng, Huang, Kreis, Fidler, Liu, and Lin}]{lin2023magic3d}
Lin, C.-H.; Gao, J.; Tang, L.; Takikawa, T.; Zeng, X.; Huang, X.; Kreis, K.; Fidler, S.; Liu, M.-Y.; and Lin, T.-Y. 2023.
\newblock Magic3d: High-resolution text-to-3d content creation.
\newblock In \emph{Proceedings of the IEEE/CVF Conference on Computer Vision and Pattern Recognition}, 300--309.

\bibitem[{Liu et~al.(2024)Liu, Xu, Jin, Chen, Varma~T, Xu, and Su}]{liu2024one}
Liu, M.; Xu, C.; Jin, H.; Chen, L.; Varma~T, M.; Xu, Z.; and Su, H. 2024.
\newblock One-2-3-45: Any single image to 3d mesh in 45 seconds without per-shape optimization.
\newblock \emph{Advances in Neural Information Processing Systems}, 36.

\bibitem[{Liu et~al.(2023{\natexlab{a}})Liu, Wu, Van~Hoorick, Tokmakov, Zakharov, and Vondrick}]{liu2023zero}
Liu, R.; Wu, R.; Van~Hoorick, B.; Tokmakov, P.; Zakharov, S.; and Vondrick, C. 2023{\natexlab{a}}.
\newblock Zero-1-to-3: Zero-shot one image to 3d object.
\newblock In \emph{Proceedings of the IEEE/CVF International Conference on Computer Vision}, 9298--9309.

\bibitem[{Liu et~al.(2023{\natexlab{b}})Liu, Lin, Zeng, Long, Liu, Komura, and Wang}]{liu2023syncdreamer}
Liu, Y.; Lin, C.; Zeng, Z.; Long, X.; Liu, L.; Komura, T.; and Wang, W. 2023{\natexlab{b}}.
\newblock Syncdreamer: Generating multiview-consistent images from a single-view image.
\newblock \emph{arXiv preprint arXiv:2309.03453}.

\bibitem[{Liu et~al.(2023{\natexlab{c}})Liu, Xie, Liu, and Wong}]{liu2023text}
Liu, Y.; Xie, M.; Liu, H.; and Wong, T.-T. 2023{\natexlab{c}}.
\newblock Text-Guided Texturing by Synchronized Multi-View Diffusion.
\newblock \emph{arXiv preprint arXiv:2311.12891}.

\bibitem[{Long et~al.(2023)Long, Guo, Lin, Liu, Dou, Liu, Ma, Zhang, Habermann, Theobalt et~al.}]{long2023wonder3d}
Long, X.; Guo, Y.-C.; Lin, C.; Liu, Y.; Dou, Z.; Liu, L.; Ma, Y.; Zhang, S.-H.; Habermann, M.; Theobalt, C.; et~al. 2023.
\newblock Wonder3d: Single image to 3d using cross-domain diffusion.
\newblock \emph{arXiv preprint arXiv:2310.15008}.

\bibitem[{Meshy(2024)}]{meshy}
Meshy. 2024.
\newblock Meshy | 3D AI Generator.
\newblock \url{https://www.meshy.ai/}.

\bibitem[{Metzer et~al.(2023)Metzer, Richardson, Patashnik, Giryes, and Cohen-Or}]{metzer2023latent}
Metzer, G.; Richardson, E.; Patashnik, O.; Giryes, R.; and Cohen-Or, D. 2023.
\newblock Latent-nerf for shape-guided generation of 3d shapes and textures.
\newblock In \emph{Proceedings of the IEEE/CVF Conference on Computer Vision and Pattern Recognition}, 12663--12673.

\bibitem[{Mou et~al.(2024)Mou, Wang, Xie, Wu, Zhang, Qi, and Shan}]{mou2024t2i}
Mou, C.; Wang, X.; Xie, L.; Wu, Y.; Zhang, J.; Qi, Z.; and Shan, Y. 2024.
\newblock T2i-adapter: Learning adapters to dig out more controllable ability for text-to-image diffusion models.
\newblock In \emph{Proceedings of the AAAI Conference on Artificial Intelligence}, volume 38(5), 4296--4304.

\bibitem[{Nichol et~al.(2021)Nichol, Dhariwal, Ramesh, Shyam, Mishkin, McGrew, Sutskever, and Chen}]{nichol2021glide}
Nichol, A.; Dhariwal, P.; Ramesh, A.; Shyam, P.; Mishkin, P.; McGrew, B.; Sutskever, I.; and Chen, M. 2021.
\newblock Glide: Towards photorealistic image generation and editing with text-guided diffusion models.
\newblock \emph{arXiv preprint arXiv:2112.10741}.

\bibitem[{Nichol et~al.(2022)Nichol, Jun, Dhariwal, Mishkin, and Chen}]{nichol2022point}
Nichol, A.; Jun, H.; Dhariwal, P.; Mishkin, P.; and Chen, M. 2022.
\newblock Point-e: A system for generating 3d point clouds from complex prompts.
\newblock \emph{arXiv preprint arXiv:2212.08751}.

\bibitem[{Oechsle et~al.(2019)Oechsle, Mescheder, Niemeyer, Strauss, and Geiger}]{oechsle2019texture}
Oechsle, M.; Mescheder, L.; Niemeyer, M.; Strauss, T.; and Geiger, A. 2019.
\newblock Texture fields: Learning texture representations in function space.
\newblock In \emph{Proceedings of the IEEE/CVF International Conference on Computer Vision}, 4531--4540.

\bibitem[{Podell et~al.(2023)Podell, English, Lacey, Blattmann, Dockhorn, M{\"u}ller, Penna, and Rombach}]{podell2023sdxl}
Podell, D.; English, Z.; Lacey, K.; Blattmann, A.; Dockhorn, T.; M{\"u}ller, J.; Penna, J.; and Rombach, R. 2023.
\newblock Sdxl: Improving latent diffusion models for high-resolution image synthesis.
\newblock \emph{arXiv preprint arXiv:2307.01952}.

\bibitem[{Poole et~al.(2022)Poole, Jain, Barron, and Mildenhall}]{poole2022dreamfusion}
Poole, B.; Jain, A.; Barron, J.~T.; and Mildenhall, B. 2022.
\newblock Dreamfusion: Text-to-3d using 2d diffusion.
\newblock \emph{arXiv preprint arXiv:2209.14988}.

\bibitem[{Radford et~al.(2021)Radford, Kim, Hallacy, Ramesh, Goh, Agarwal, Sastry, Askell, Mishkin, Clark et~al.}]{radford2021learning}
Radford, A.; Kim, J.~W.; Hallacy, C.; Ramesh, A.; Goh, G.; Agarwal, S.; Sastry, G.; Askell, A.; Mishkin, P.; Clark, J.; et~al. 2021.
\newblock Learning transferable visual models from natural language supervision.
\newblock In \emph{International conference on machine learning}, 8748--8763. PMLR.

\bibitem[{Ramesh et~al.(2022)Ramesh, Dhariwal, Nichol, Chu, and Chen}]{ramesh2022hierarchical}
Ramesh, A.; Dhariwal, P.; Nichol, A.; Chu, C.; and Chen, M. 2022.
\newblock Hierarchical text-conditional image generation with clip latents.
\newblock \emph{arXiv preprint arXiv:2204.06125}, 1(2): 3.

\bibitem[{Richardson et~al.(2023)Richardson, Metzer, Alaluf, Giryes, and Cohen-Or}]{richardson2023texture}
Richardson, E.; Metzer, G.; Alaluf, Y.; Giryes, R.; and Cohen-Or, D. 2023.
\newblock Texture: Text-guided texturing of 3d shapes.
\newblock In \emph{ACM SIGGRAPH 2023 Conference Proceedings}, 1--11.

\bibitem[{Rombach et~al.(2022)Rombach, Blattmann, Lorenz, Esser, and Ommer}]{rombach2022high}
Rombach, R.; Blattmann, A.; Lorenz, D.; Esser, P.; and Ommer, B. 2022.
\newblock High-resolution image synthesis with latent diffusion models.
\newblock In \emph{Proceedings of the IEEE/CVF conference on computer vision and pattern recognition}, 10684--10695.

\bibitem[{Saharia et~al.(2022)Saharia, Chan, Saxena, Li, Whang, Denton, Ghasemipour, Gontijo~Lopes, Karagol~Ayan, Salimans et~al.}]{saharia2022photorealistic}
Saharia, C.; Chan, W.; Saxena, S.; Li, L.; Whang, J.; Denton, E.~L.; Ghasemipour, K.; Gontijo~Lopes, R.; Karagol~Ayan, B.; Salimans, T.; et~al. 2022.
\newblock Photorealistic text-to-image diffusion models with deep language understanding.
\newblock \emph{Advances in neural information processing systems}, 35: 36479--36494.

\bibitem[{Schuhmann et~al.(2022)Schuhmann, Beaumont, Vencu, Gordon, Wightman, Cherti, Coombes, Katta, Mullis, Wortsman et~al.}]{schuhmann2022laion}
Schuhmann, C.; Beaumont, R.; Vencu, R.; Gordon, C.; Wightman, R.; Cherti, M.; Coombes, T.; Katta, A.; Mullis, C.; Wortsman, M.; et~al. 2022.
\newblock Laion-5b: An open large-scale dataset for training next generation image-text models.
\newblock \emph{Advances in Neural Information Processing Systems}, 35: 25278--25294.

\bibitem[{Shi et~al.(2023)Shi, Wang, Ye, Long, Li, and Yang}]{shi2023mvdream}
Shi, Y.; Wang, P.; Ye, J.; Long, M.; Li, K.; and Yang, X. 2023.
\newblock Mvdream: Multi-view diffusion for 3d generation.
\newblock \emph{arXiv preprint arXiv:2308.16512}.

\bibitem[{Siddiqui et~al.(2022)Siddiqui, Thies, Ma, Shan, Nie{\ss}ner, and Dai}]{siddiqui2022texturify}
Siddiqui, Y.; Thies, J.; Ma, F.; Shan, Q.; Nie{\ss}ner, M.; and Dai, A. 2022.
\newblock Texturify: Generating textures on 3d shape surfaces.
\newblock In \emph{European Conference on Computer Vision}, 72--88. Springer.

\bibitem[{Tsalicoglou et~al.(2023)Tsalicoglou, Manhardt, Tonioni, Niemeyer, and Tombari}]{tsalicoglou2023textmesh}
Tsalicoglou, C.; Manhardt, F.; Tonioni, A.; Niemeyer, M.; and Tombari, F. 2023.
\newblock Textmesh: Generation of realistic 3d meshes from text prompts.
\newblock \emph{arXiv preprint arXiv:2304.12439}.

\bibitem[{Turk and Levoy(1994)}]{turk1994zippered}
Turk, G.; and Levoy, M. 1994.
\newblock Zippered polygon meshes from range images.
\newblock In \emph{Proceedings of the 21st annual conference on Computer graphics and interactive techniques}, 311--318.

\bibitem[{Wang et~al.(2023)Wang, Du, Li, Yeh, and Shakhnarovich}]{wang2023score}
Wang, H.; Du, X.; Li, J.; Yeh, R.~A.; and Shakhnarovich, G. 2023.
\newblock Score jacobian chaining: Lifting pretrained 2d diffusion models for 3d generation.
\newblock In \emph{Proceedings of the IEEE/CVF Conference on Computer Vision and Pattern Recognition}, 12619--12629.

\bibitem[{Wang et~al.(2024)Wang, Lu, Wang, Bao, Li, Su, and Zhu}]{wang2024prolificdreamer}
Wang, Z.; Lu, C.; Wang, Y.; Bao, F.; Li, C.; Su, H.; and Zhu, J. 2024.
\newblock Prolificdreamer: High-fidelity and diverse text-to-3d generation with variational score distillation.
\newblock \emph{Advances in Neural Information Processing Systems}, 36.

\bibitem[{Xu et~al.(2023)Xu, Tan, Luan, Bi, Wang, Li, Shi, Sunkavalli, Wetzstein, Xu et~al.}]{xu2023dmv3d}
Xu, Y.; Tan, H.; Luan, F.; Bi, S.; Wang, P.; Li, J.; Shi, Z.; Sunkavalli, K.; Wetzstein, G.; Xu, Z.; et~al. 2023.
\newblock Dmv3d: Denoising multi-view diffusion using 3d large reconstruction model.
\newblock \emph{arXiv preprint arXiv:2311.09217}.

\bibitem[{Yang et~al.(2023)Yang, Zhou, Liu, , and Loy}]{yang2023rerender}
Yang, S.; Zhou, Y.; Liu, Z.; ; and Loy, C.~C. 2023.
\newblock Rerender A Video: Zero-Shot Text-Guided Video-to-Video Translation.
\newblock In \emph{ACM SIGGRAPH Asia Conference Proceedings}.

\bibitem[{Yang et~al.(2024)Yang, Zhou, Liu, and Loy}]{yang2024fresco}
Yang, S.; Zhou, Y.; Liu, Z.; and Loy, C.~C. 2024.
\newblock FRESCO: Spatial-Temporal Correspondence for Zero-Shot Video Translation.
\newblock \emph{arXiv preprint arXiv:2403.12962}.

\bibitem[{Young(2016)}]{xatlas2016}
Young, J. 2016.
\newblock xatlas.
\newblock In \emph{github.com/jpcy/xatlas}.

\bibitem[{Youwang, Oh, and Pons-Moll(2023)}]{youwang2023paint}
Youwang, K.; Oh, T.-H.; and Pons-Moll, G. 2023.
\newblock Paint-it: Text-to-Texture Synthesis via Deep Convolutional Texture Map Optimization and Physically-Based Rendering.
\newblock \emph{arXiv preprint arXiv:2312.11360}.

\bibitem[{Yu et~al.(2021)Yu, Dong, Peers, and Tong}]{yu2021learning}
Yu, R.; Dong, Y.; Peers, P.; and Tong, X. 2021.
\newblock Learning texture generators for 3d shape collections from internet photo sets.
\newblock In \emph{British Machine Vision Conference}.

\bibitem[{Zhang, Rao, and Agrawala(2023)}]{zhang2023adding}
Zhang, L.; Rao, A.; and Agrawala, M. 2023.
\newblock Adding conditional control to text-to-image diffusion models.
\newblock In \emph{Proceedings of the IEEE/CVF International Conference on Computer Vision}, 3836--3847.

\end{thebibliography}

\newpage
\clearpage
\appendix

\section{Surface space color dilation}
\begin{figure}
  \centering
  \includegraphics[width=0.9\linewidth]{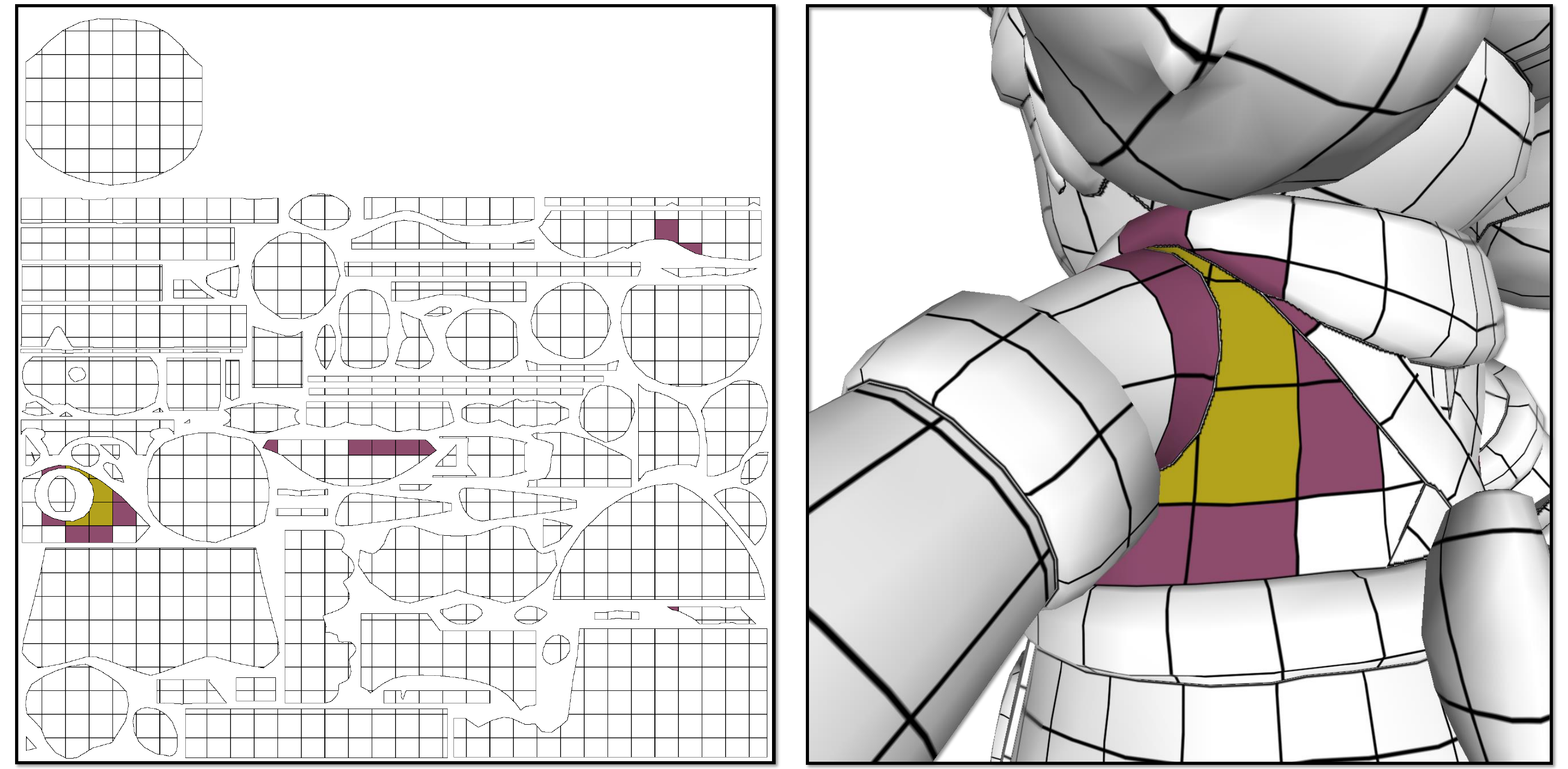}
  \caption{
    An illustration on our texture dilation algorithm. The yellow area can be influenced by the neighbor regions in surface space. Note how the colors can be propagate between distant UV islands.  
  }
  \label{fig:dilation}

\end{figure}

\begin{figure}
  \centering
  \includegraphics[width=\linewidth]{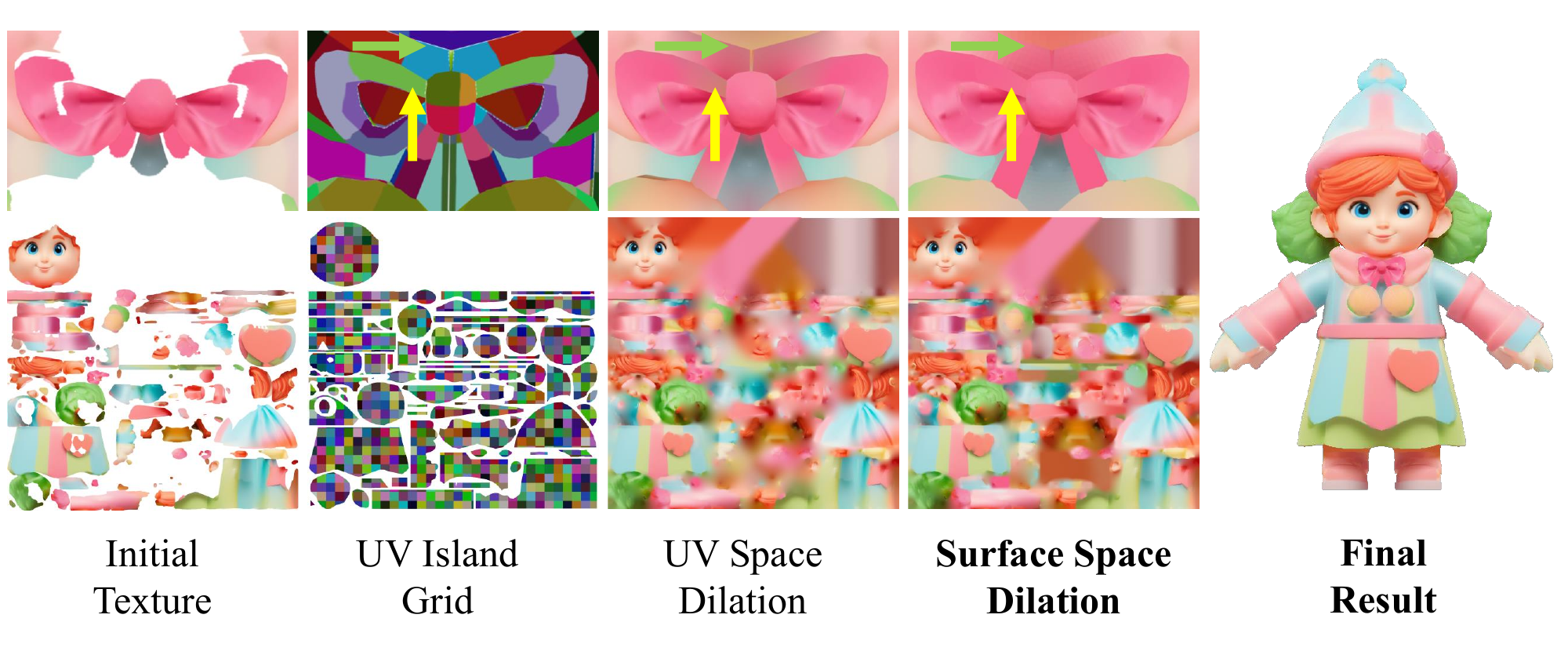}
  \caption{
An illustration of surface space color propagation algorithm for texture completion. Our method propagates valid texture color in surface space instead of UV space. This effectively addresses inaccurate color propagation when two points are proximate to each other in 3D but situated on remote UV islands (green arrow), or located on nearby UV islands but having a large 3D distance (yellow arrow).    
  }
  \label{fig:ablation_surface_dilation}
\end{figure}

We first divide the original UV map into sub-UV islands using equal-sized grids, as illustrated in Fig.~\ref{fig:dilation} and Fig.~\ref{fig:ablation_surface_dilation}. Next, we calculate the connectivity of sub-UV islands and generate an adjacency matrix. Then, we iteratively traverse the invalid pixels on the UV map which are invisible from all perspectives. For each invalid pixel, we first pick candidates from textured pixels based on their relative distance in 3D, the cosine similarity of their vertex normal, and the connectivity recorded by the adjacency matrix. We then calculate the color for the invalid pixel by performing a weighted average of these candidates. We iterate this algorithm until all invalid pixels are filled or reach the max step. The detailed algorithm on surface space color dilation is shown in Algorithm.~\ref{algorithm:surface_dilation}. An illustration of this process is shown in Fig.~\ref{fig:dilation}. As demonstrated in column 3 of Fig.~\ref{fig:ablation_surface_dilation}, the UV space dilation method may propagate colors from pixels that are not adjacent in the 3D space, resulting in inaccuracies in the final texture map. In contrast, our surface space color dilation algorithm propagates valid texture color in surface space instead of UV space, thereby effectively addresses inaccurate color propagation when using naive flood-fill method in UV space. \\
\section{Implementation details}
We implement our algorithm using an open-source framework: ComfyUI\cite{ComfyUI}, and we adopt nvdiffrast~\cite{Laine2020diffrast} for rendering and inverse rendering. We set the strength of ControlNet as $1.0$ in all our experiments. As for parameters of surface space dilation algorithm, the grid size $s=64$, the distance threshold $d_{th}=0.02$, the angle threshold $a_{th}=90^{\circ}$, the nearest neighbors number $n=30$, and iterations $iter=10$.

\begin{algorithm}
\caption{UV dilation in surface space}
\label{algorithm:surface_dilation}
\SetAlgoNoLine
\KwIn{\\
input UV map $U$\\
uv-space spatial position map $X$\\
uv-space normal map $N$\\
uv-space face index map $F$\\
uv-space visibility map $M$\\

\\
}
\textbf{Parameters:} grid size $s$, dilation distance threshold $d_{th}$, dilation angle threshold $a_{th}$, iterations $iter$, number of nearest neighbors $n$\\
\KwOut{UV map after dilation $U$}
~\\
$I_{ori} \gets get\_original\_uv\_island(F)$\\
$I_{grid} \gets get\_grid\_uv\_island(F, s)$\\
$M_{adj} \gets get\_adjacency\_matrix(F, I_{grid},I_{ori})$\\
$P, Q \gets get\_valid\_invalid\_points(M)  $\\
\For{$i = 1, 2, \dots, iter$}{
    \For{each $q \in Q$}{
        $A = I_{grid}[q]$\\
        $q_{n} \gets KNN(q, P, n)$\\
        \For{each $q_{k} \in q_{n}$}{
            $B = I_{grid}[q_{k}]$\\
            $dist = ||X[q] - X[q_{k}]||_2$\\
            $angle = angle\_between(N[q], N[q_k])$\\
            \eIf{$q_{k} \notin Q$ \textbf{and} $ angle < a_{th}$ \textbf{and} $M_{adj}[A][B] == True$ \textbf{and} $dist < d_{th}$}{
                $w_{k} = 1 - (dist / d_{th})^2$
            }{
                $w_{k} = 0$
            }
        }
        $w = \sum_{q_k \in q_n} w_{k}$\\
        \If{$w \neq 0$}{
            $U[q] = \frac{1}{w}\sum_{q_k \in q_n}(U[q_k] * w_{k}) $\\
            remove $q$ from $Q$
        }
    }
}

\end{algorithm}


\section{More Results}
\begin{figure}
  \centering
  \includegraphics[width=\linewidth]{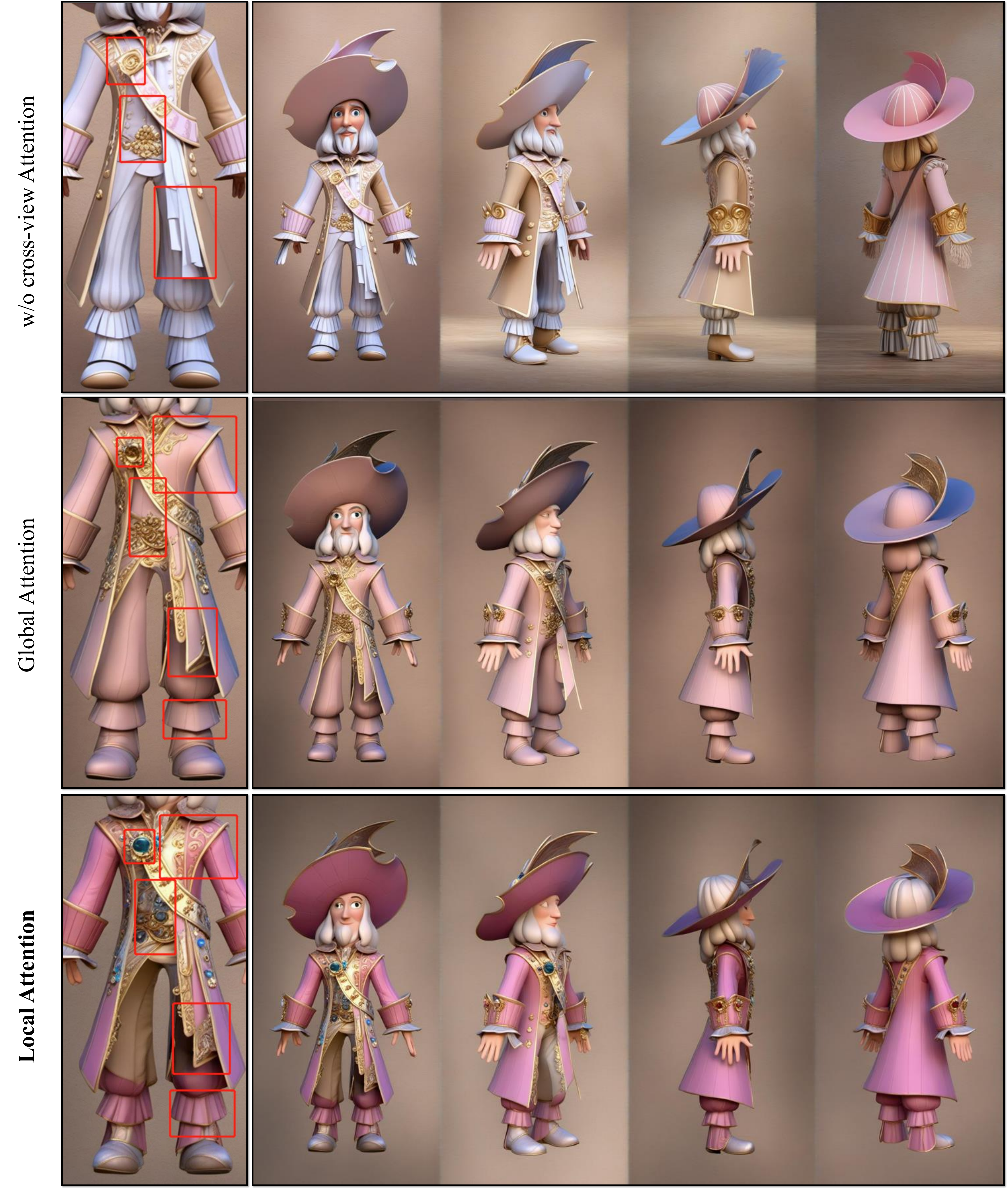}
  \caption{
    Ablation results on different attention mechanisms in multi-view generation. Each view attends to its neighbors (top), each view attends to all other views (middle), our \textbf{local attention} (bottom) achieves the best multi-view consistency while preserving rich details.
  }
  \label{fig:ablation_local_attn_2}
\end{figure}

\begin{figure}
  \centering
  \includegraphics[width=\linewidth]{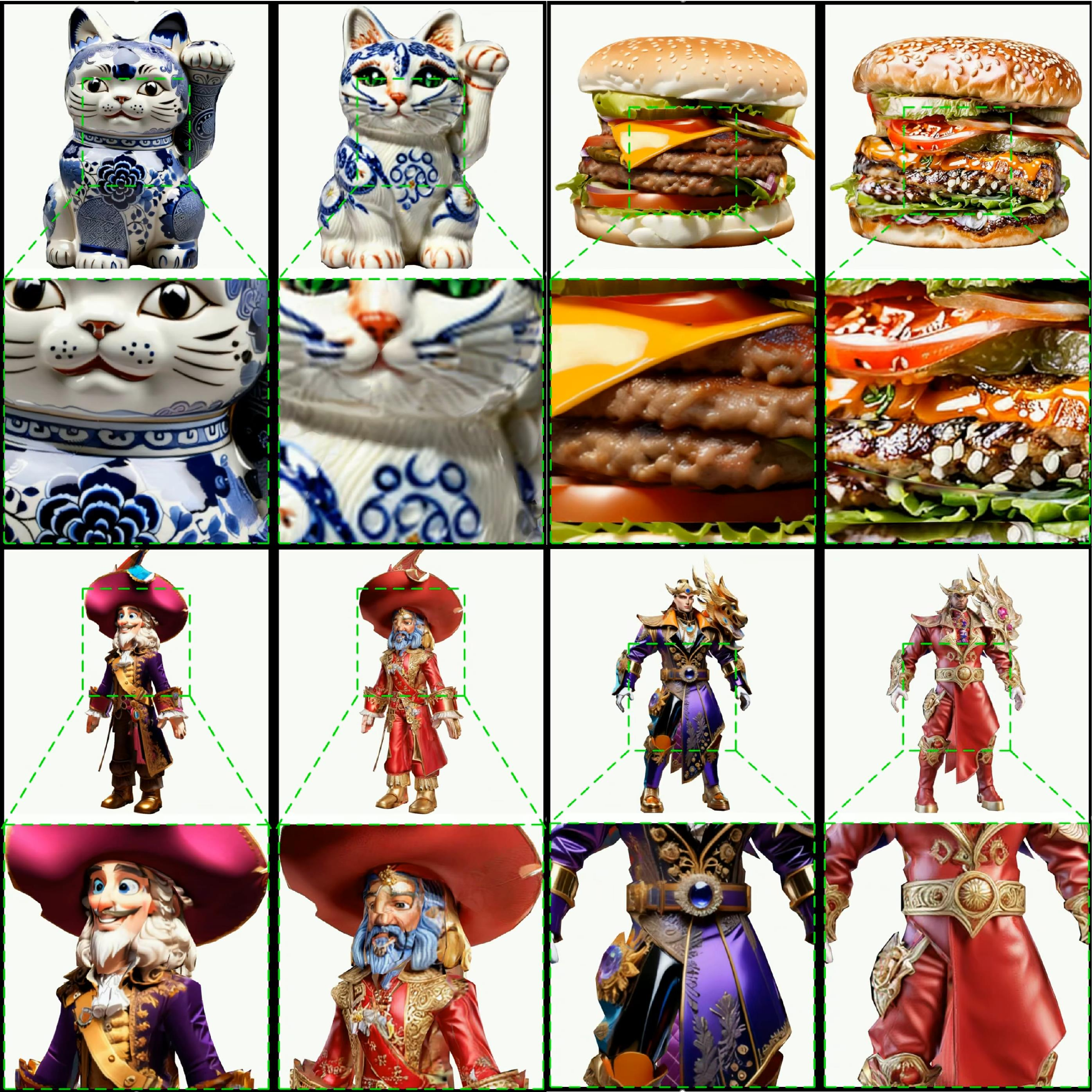}
  \caption{
    Qualitative comparison with Meshy-3. Our results are shown on the left for each group with Meshy-3 on the right.
  }
  \label{fig:comp_meshy}
\end{figure}

\begin{figure*}
  \centering
  \includegraphics[width=0.9\linewidth]{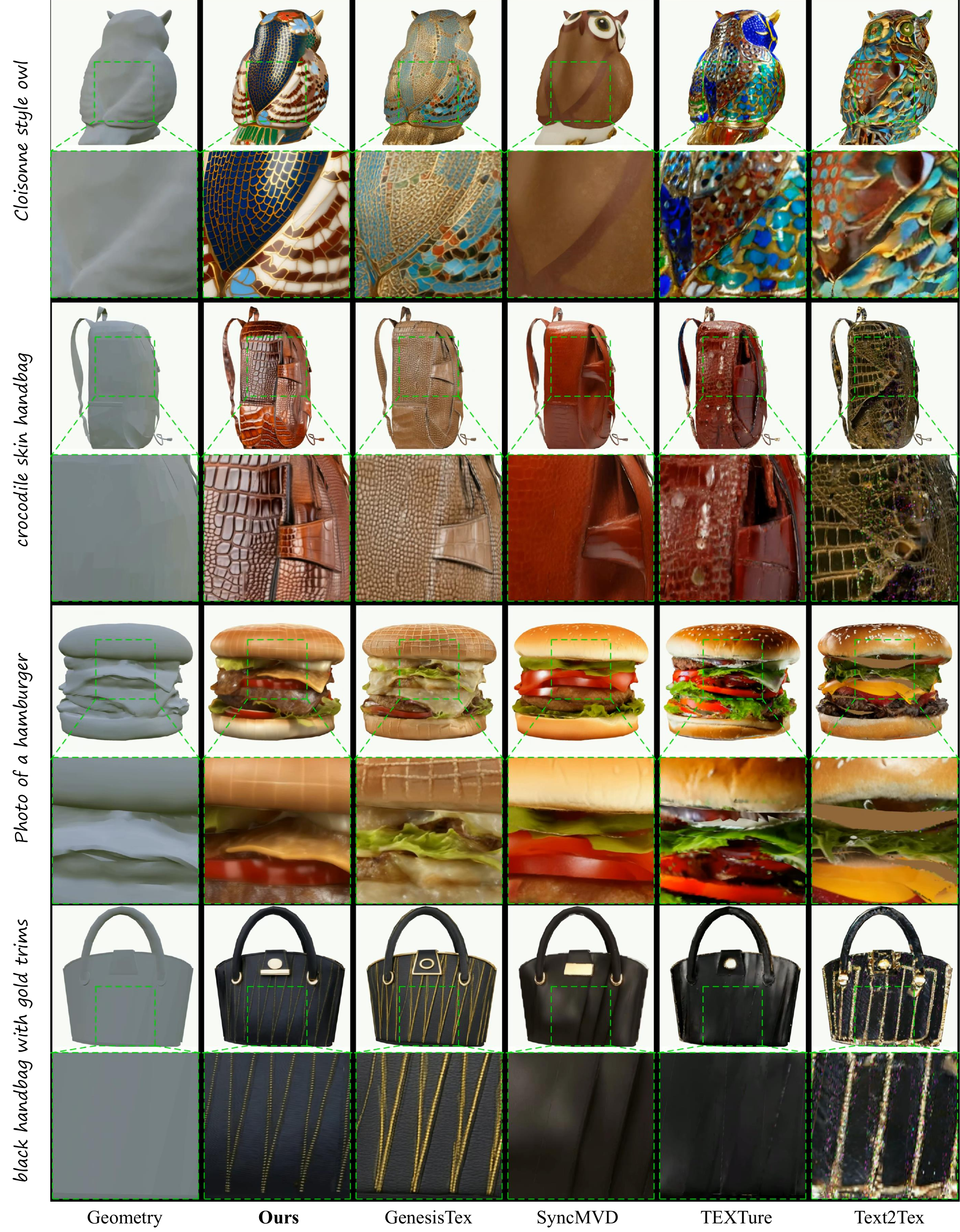}
  \caption{
    More comparison results with different methods.
  }
  \label{fig:comparison3}
\end{figure*}

\begin{figure*}
  \centering
  \includegraphics[width=0.9\linewidth]{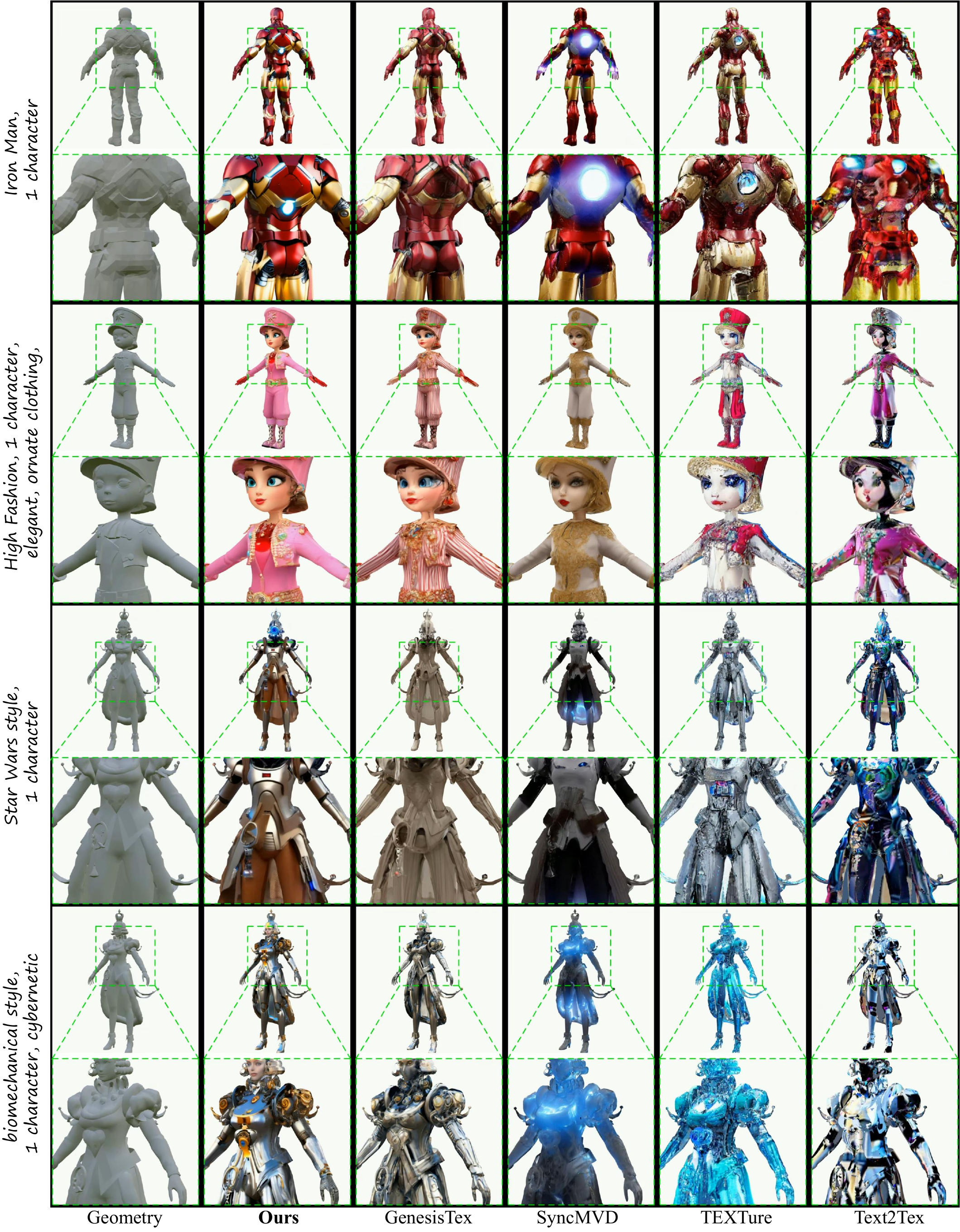}
  \caption{
    More comparison results with different methods.
  }
  \label{fig:comparison4}
\end{figure*}

\begin{figure*}
  \centering
  \includegraphics[width=0.9\linewidth]{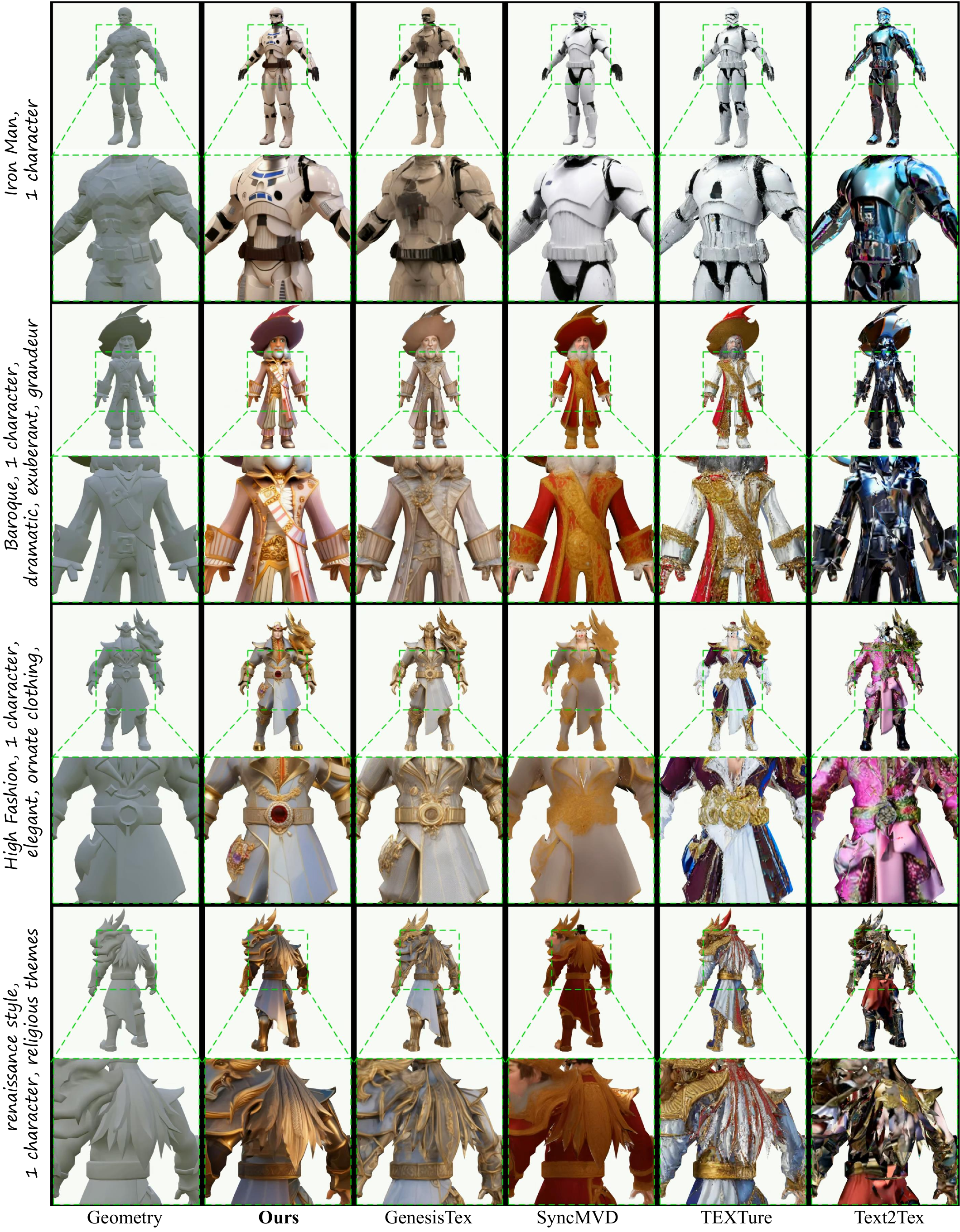}
  \caption{
    More comparison results with different methods.
  }
  \label{fig:comparison5}
\end{figure*}

\begin{figure*}
  \centering
  \includegraphics[width=\linewidth]{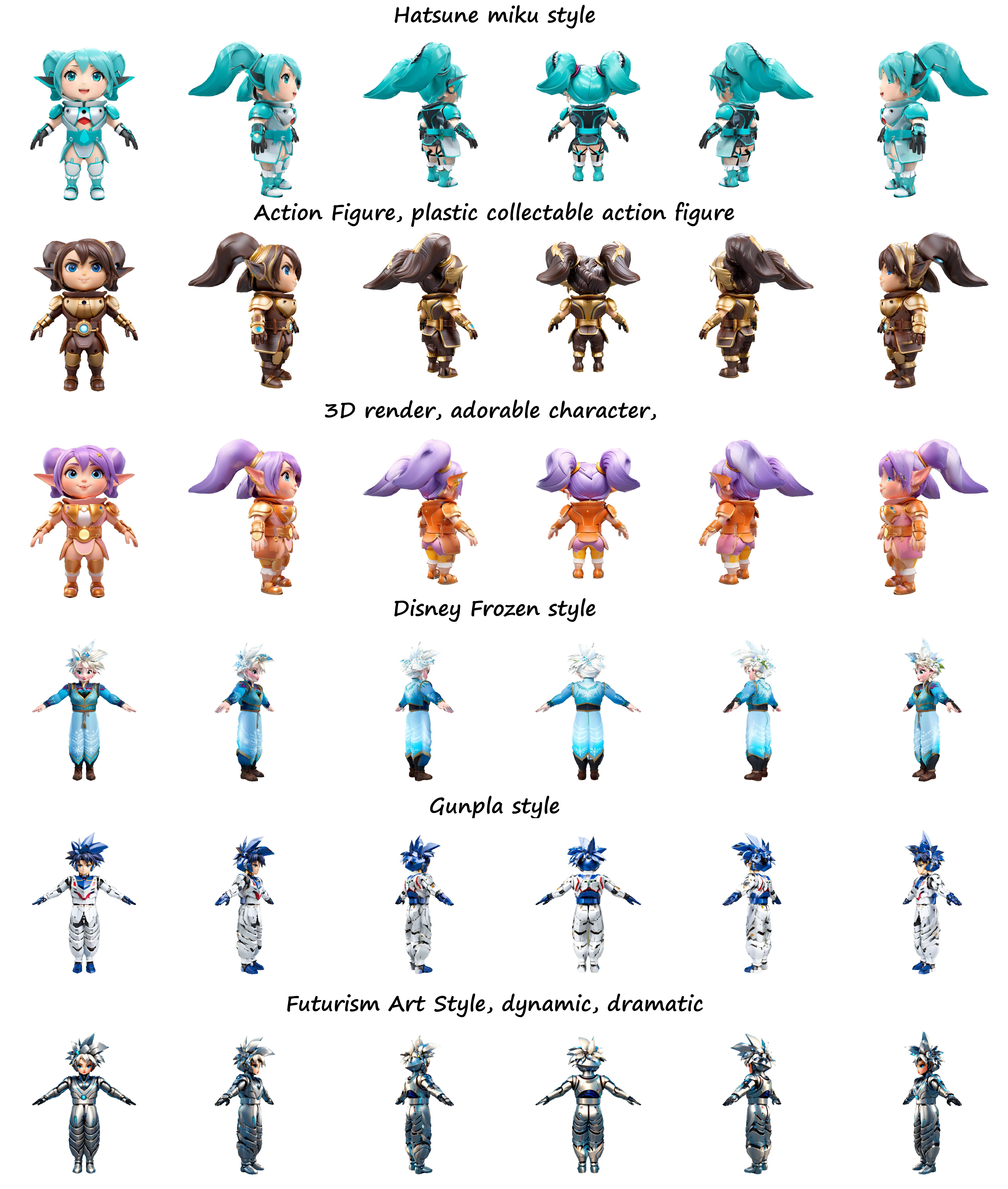}
  \caption{
    More results on meshes from objaverse\cite{deitke2023objaverse}.
  }
  \label{fig:more_obj0}
\end{figure*}

\begin{figure*}
  \centering
  \includegraphics[width=\linewidth]{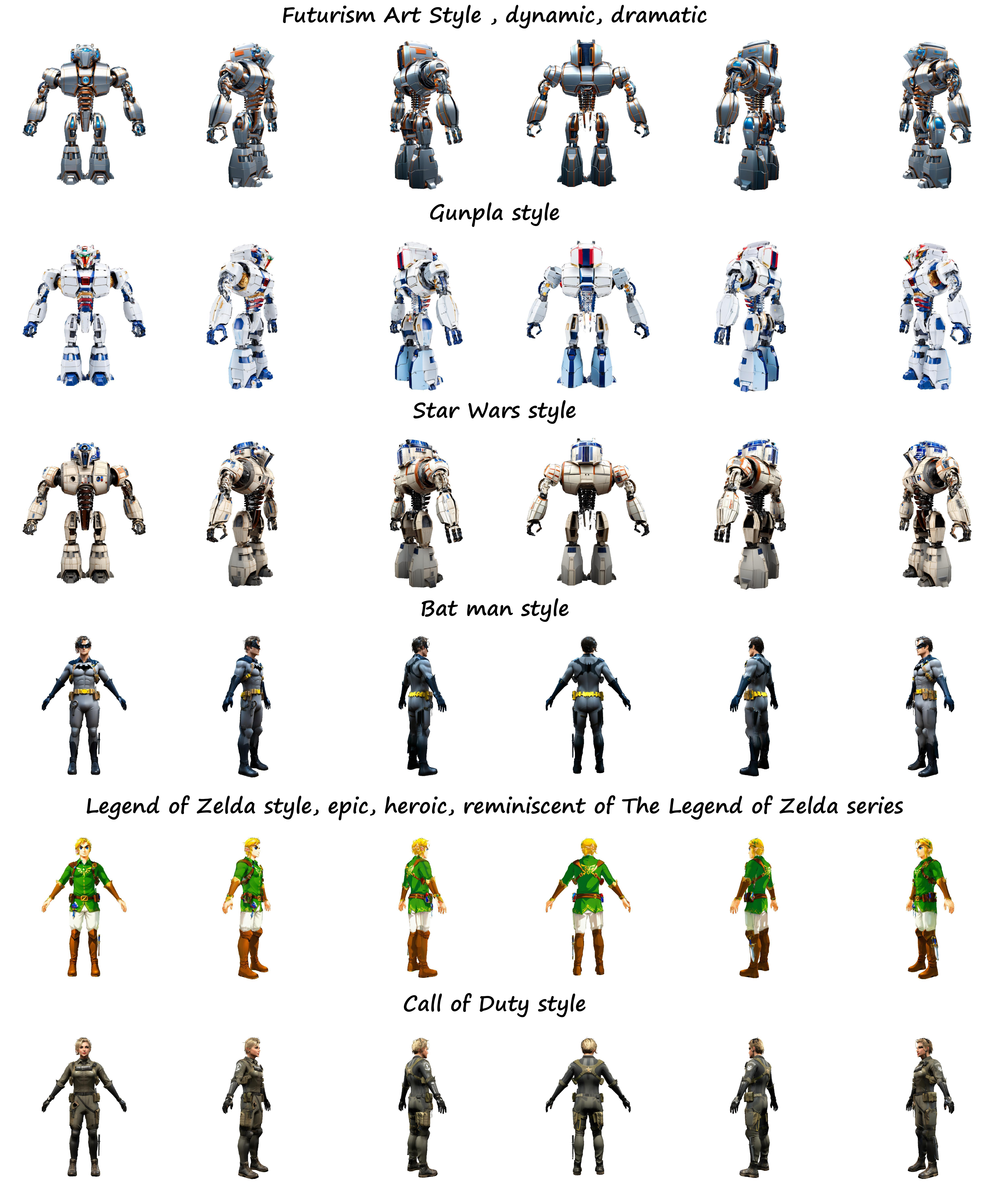}
  \caption{
    More results on meshes from objaverse\cite{deitke2023objaverse}. 
  }
  \label{fig:more_obj1}
\end{figure*}

\begin{figure*}
  \centering
  \includegraphics[width=\linewidth]{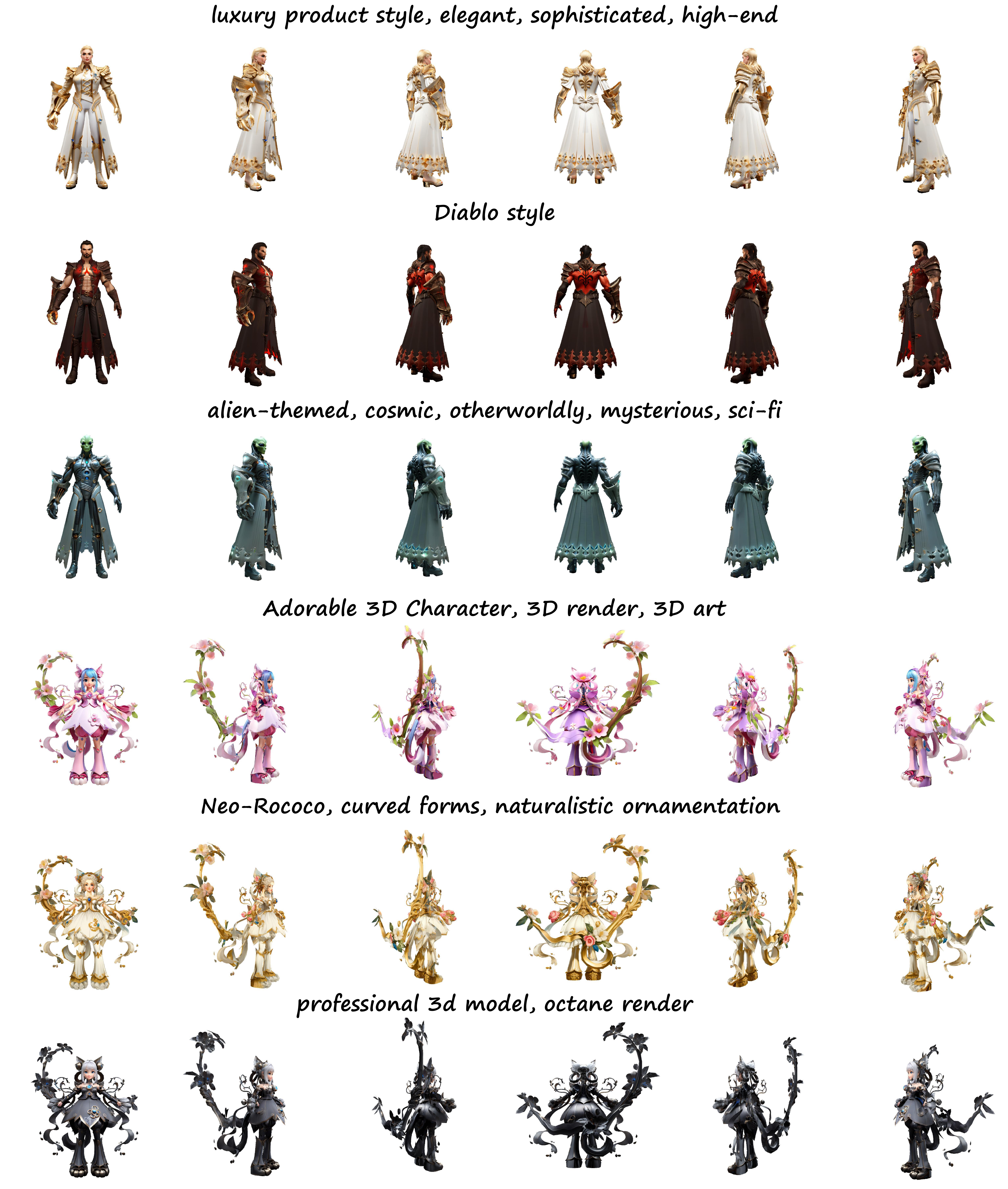}
  \caption{
    More results on meshes from objaverse\cite{deitke2023objaverse}. 
  }
  \label{fig:more_obj2}
\end{figure*}

\begin{figure*}
  \centering
  \includegraphics[width=\linewidth]{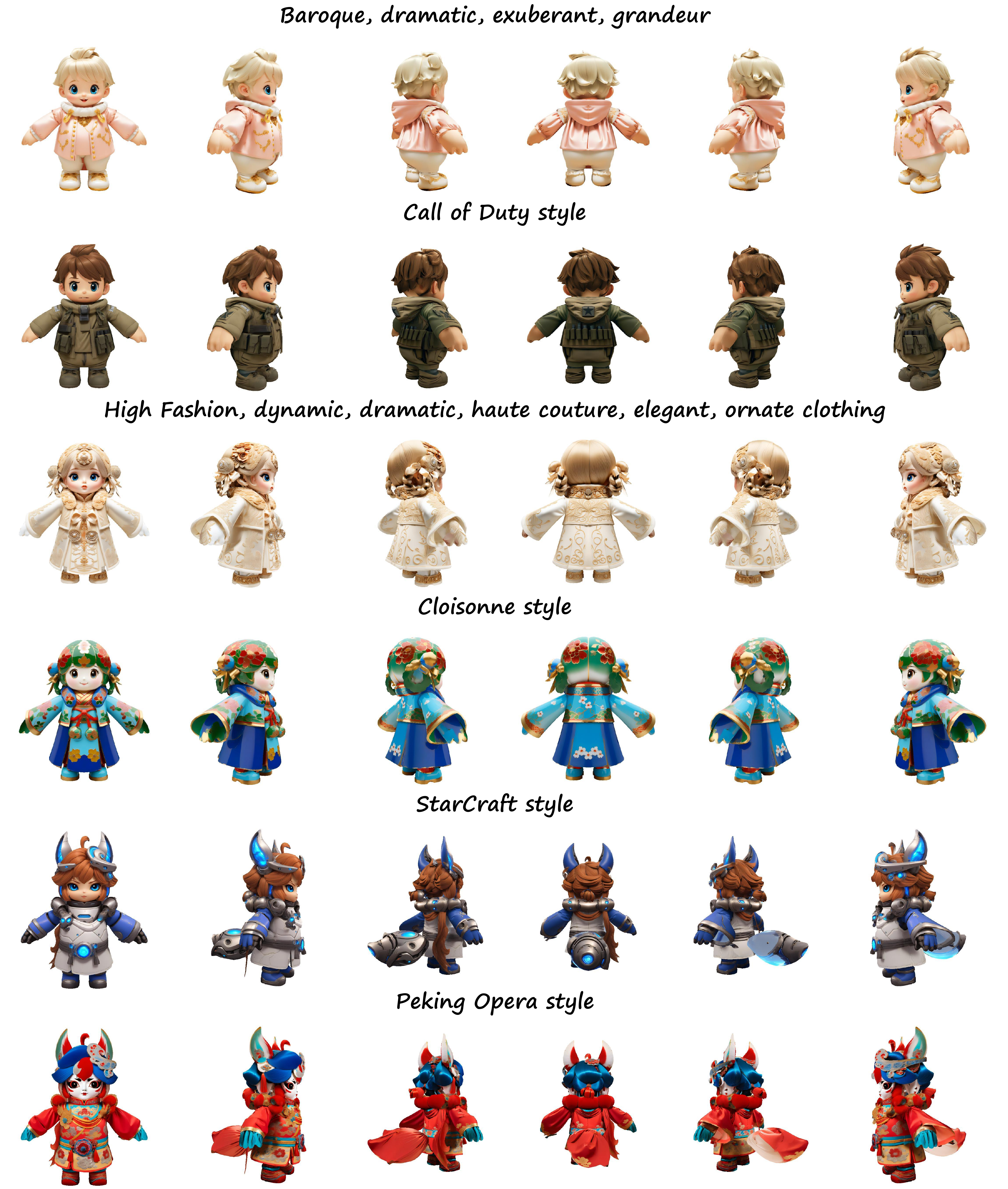}
  \caption{
    More results on meshes from industrial games. 
  }
  \label{fig:more1}
\end{figure*}

\begin{figure*}
  \centering
  \includegraphics[width=\linewidth]{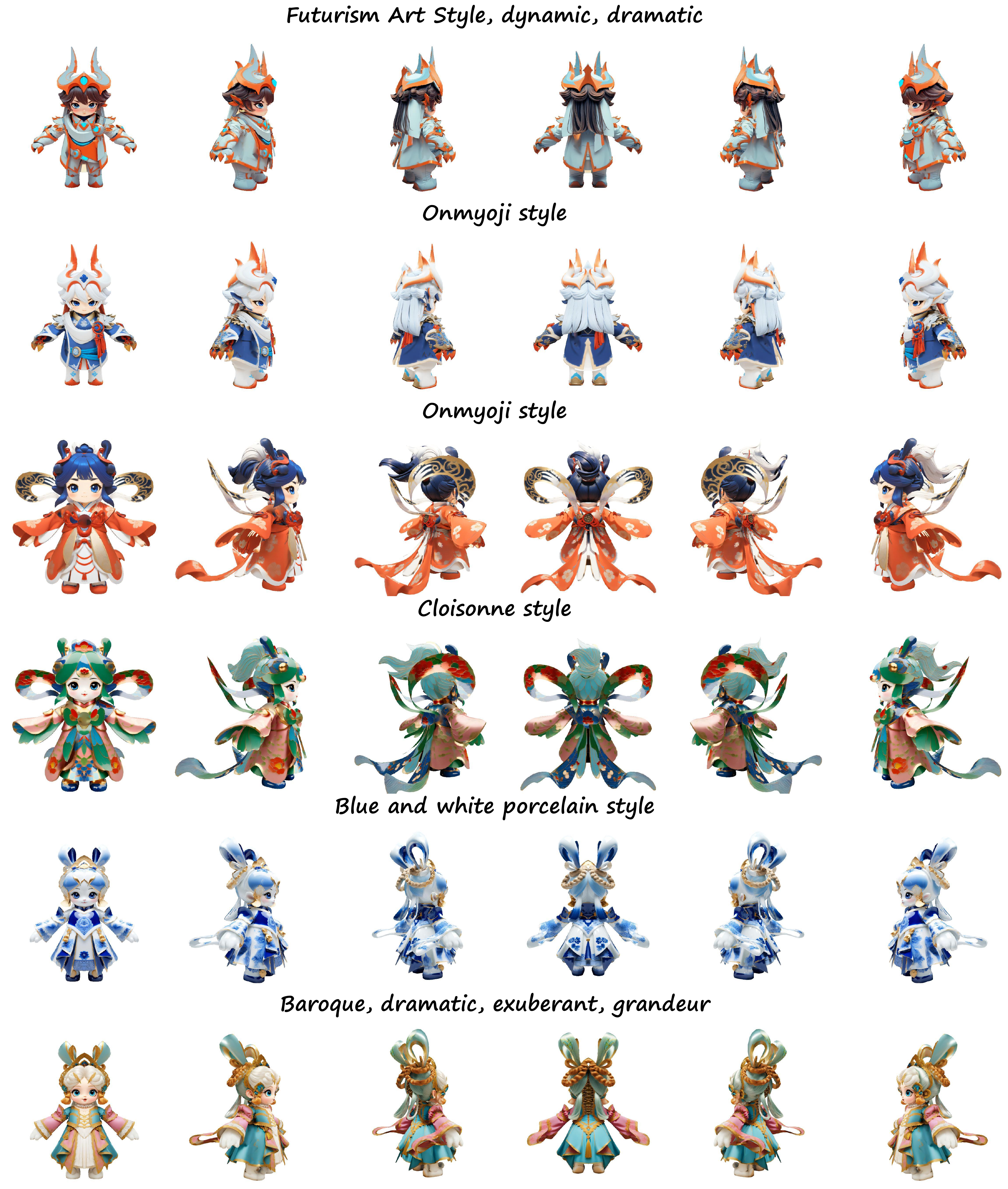}
  \caption{
    More results on meshes from industrial games.
  }
  \label{fig:more2}
\end{figure*}

\begin{figure*}
  \centering
  \includegraphics[width=\linewidth]{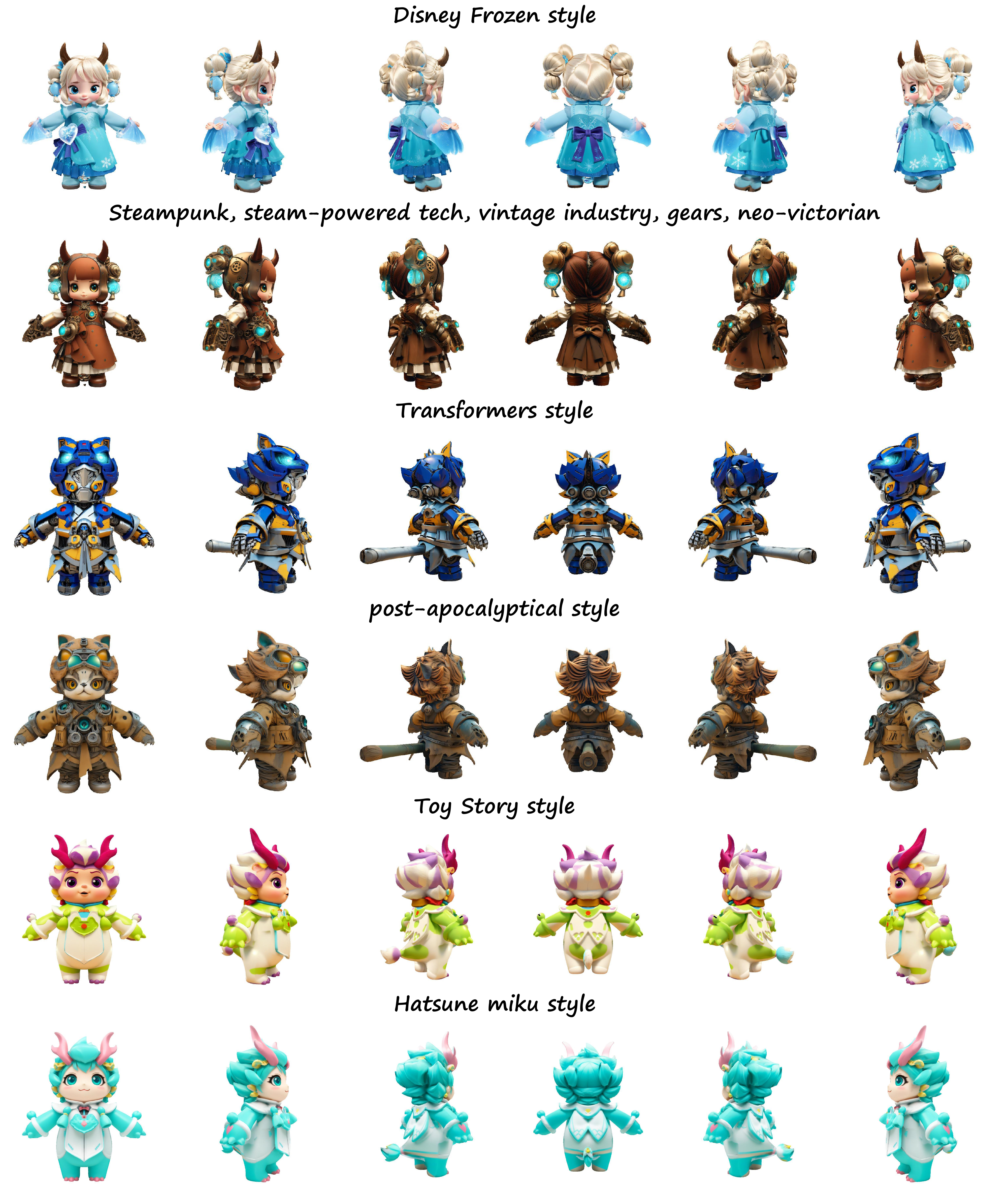}
  \caption{
    More results on meshes from industrial games.
  }
  \label{fig:more3}
\end{figure*}

We present additional ablation experiments on local attention in Fig.~\ref{fig:ablation_local_attn_2}.  This figure illustrates the ablation results for various attention mechanisms in multi-view generation without latent merging. Our local attention method demonstrates superior multi-view consistency while effectively preserving intricate details that close to the images generated by the original unconstrained diffusion (row 1).
Furthermore, we include results compared with different methods in Fig.~\ref{fig:comparison3},~\ref{fig:comparison4}, and~\ref{fig:comparison5}. The qualitative comparison with Meshy-3~\cite{meshy} can
be found in Fig.~\ref{fig:comp_meshy}. Meshy-3 produces highly contrasting colors with
considerable details but tends to generate ghosting artifacts and
sometimes over-saturated results. In contrast, our method can produce
textures with better visual quality and considerable diversity, while keeping surface consistency. Additional results showcasing our methods across various meshes and styles can be found in Fig.~\ref{fig:more_obj0},~\ref{fig:more_obj1},~\ref{fig:more_obj2},~\ref{fig:more1},~\ref{fig:more2}, and~\ref{fig:more3}.

\clearpage
\newpage

\end{document}


\title{Supplementary Materials}

\maketitle

\begin{algorithm}
\caption{UV dilation in surface space}
\label{algorithm:surface_dilation}
\SetAlgoNoLine
\KwIn{\\
input UV map $U$\\
uv-space spatial position map $X$\\
uv-space normal map $N$\\
uv-space face index map $F$\\
uv-space visibility map $M$\\

\\
}
\textbf{Parameters:} grid size $s$, dilation distance threshold $d_{th}$, dilation angle threshold $a_{th}$, iterations $iter$, number of nearest neighbors $n$\\
\KwOut{UV map after dilation $U$}
~\\
$I_{ori} \gets get\_original\_uv\_island(F)$\\
$I_{grid} \gets get\_grid\_uv\_island(F, s)$\\
$M_{adj} \gets get\_adjacency\_matrix(F, I_{grid},I_{ori})$\\
$P, Q \gets get\_valid\_invalid\_points(M)  $\\
\For{$i = 1, 2, \dots, iter$}{
    \For{each $q \in Q$}{
        $A = I_{grid}[q]$\\
        $q_{n} \gets KNN(q, P, n)$\\
        \For{each $q_{k} \in q_{n}$}{
            $B = I_{grid}[q_{k}]$\\
            $dist = ||X[q] - X[q_{k}]||_2$\\
            $angle = angle\_between(N[q], N[q_k])$\\
            \eIf{$q_{k} \notin Q$ \textbf{and} $ angle < a_{th}$ \textbf{and} $M_{adj}[A][B] == True$ \textbf{and} $dist < d_{th}$}{
                $w_{k} = 1 - (dist / d_{th})^2$
            }{
                $w_{k} = 0$
            }
        }
        $w = \sum_{q_k \in q_n} w_{k}$\\
        \If{$w \neq 0$}{
            $U[q] = \frac{1}{w}\sum_{q_k \in q_n}(U[q_k] * w_{k}) $\\
            remove $q$ from $Q$
        }
    }
}

\end{algorithm}

\section{Implementation Details}
\begin{figure}
  \centering
  \includegraphics[width=\linewidth]{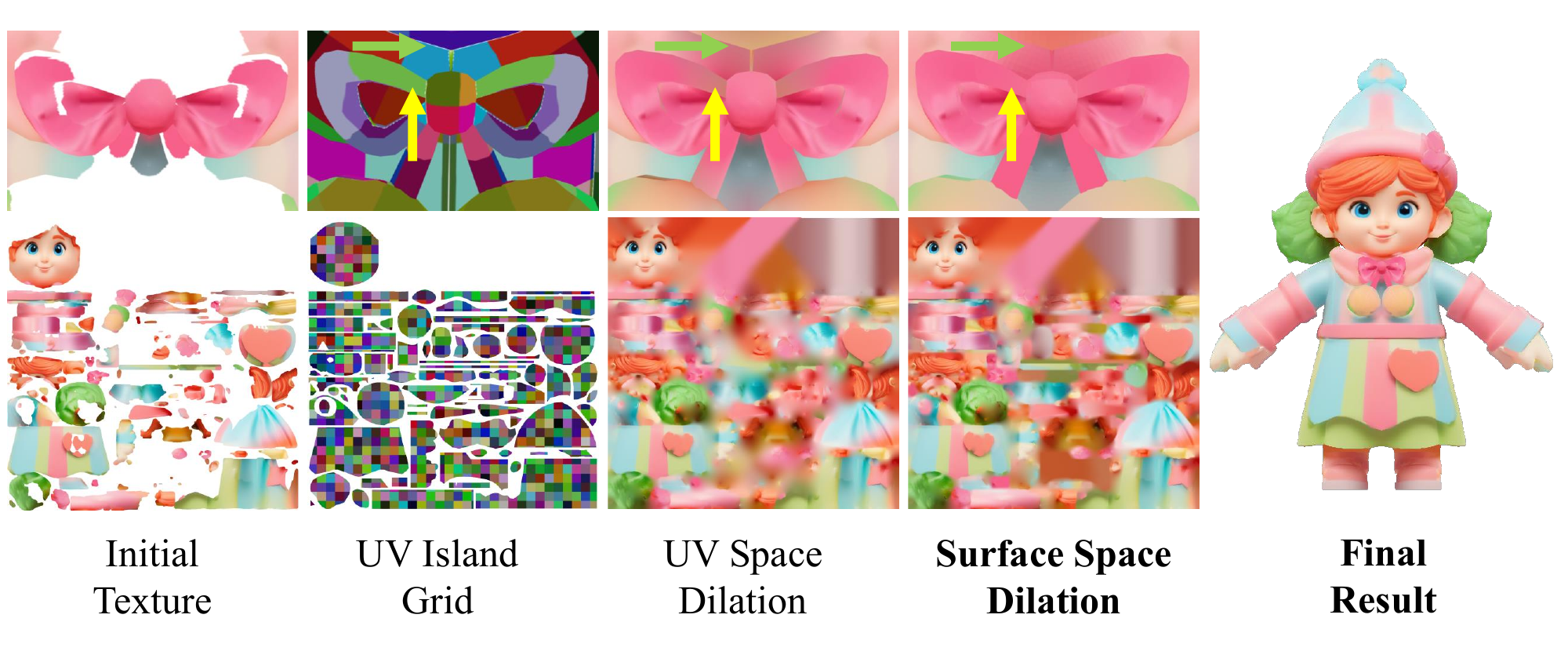}
  \caption{
An illustration of surface space color propagation algorithm for texture completion. Our method propagates valid texture color in surface space instead of UV space. This effectively addresses inaccurate color propagation when two points are proximate to each other in 3D but situated on remote UV islands (green arrow), or located on nearby UV islands but having a large 3D distance (yellow arrow).    
  }
  \label{fig:ablation_surface_dilation}
\end{figure}

\begin{figure}
  \centering
  \includegraphics[width=0.9\linewidth]{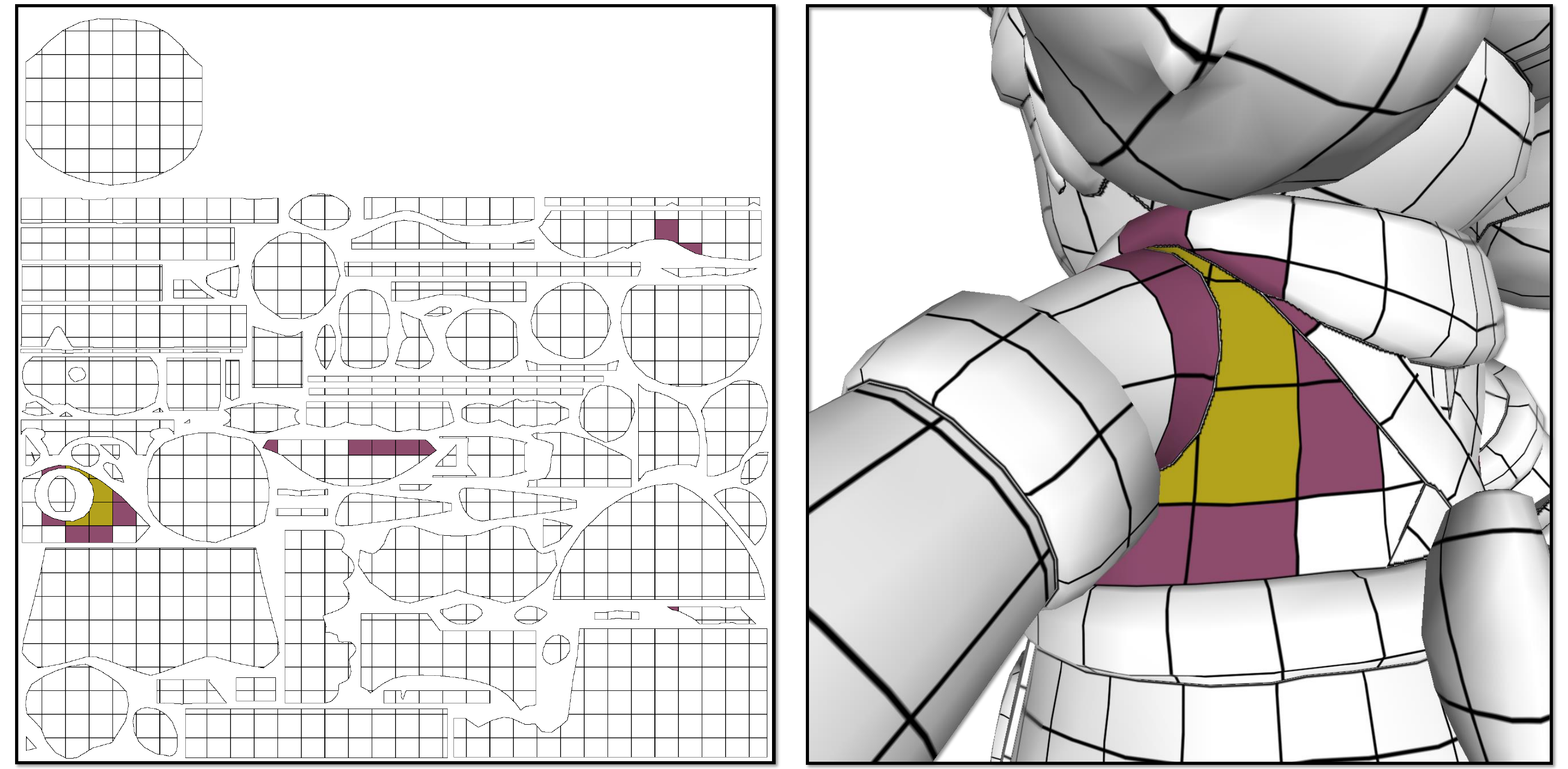}
  \caption{
    An illustration on our texture dilation algorithm. The yellow area can be influenced by the neighbor regions in surface space. Note how the colors can be propagate between distant UV islands.  
  }
  \label{fig:dilation}
\end{figure}

\textbf{Details of surface space color dilation.}
We first divide the original UV map into sub-UV islands using equal-sized grids. Next, we calculate the connectivity of sub-UV islands and generate an adjacency matrix. Then, we iteratively traverse the invalid pixels. For each invalid pixel, we first pick candidates from textured pixels based on their relative distance in 3D, the cosine similarity of their vertex normal, and the connectivity recorded by the adjacency matrix. We then calculate the color for the invalid pixel by performing a weighted average of these candidates. We iterate this algorithm until all invalid pixels are filled or reach the max step. The detailed algorithm on surface space color dilation is shown in Algorithm.~\ref{algorithm:surface_dilation}. An illustration of this process is shown in Fig.~\ref{fig:dilation}, with an ablation on the surface space color dilation shown in  Fig.~\ref{fig:ablation_surface_dilation}.\\
\textbf{More details.}
We implement our algorithm using an open-source framework: ComfyUI\cite{ComfyUI}, and we adopt nvdiffrast~\cite{Laine2020diffrast} for rendering. We set the strength of ControlNet as $1.0$ in all our experiments.


\section{More Results}
\begin{figure*}
  \centering
  \includegraphics[width=0.9\linewidth]{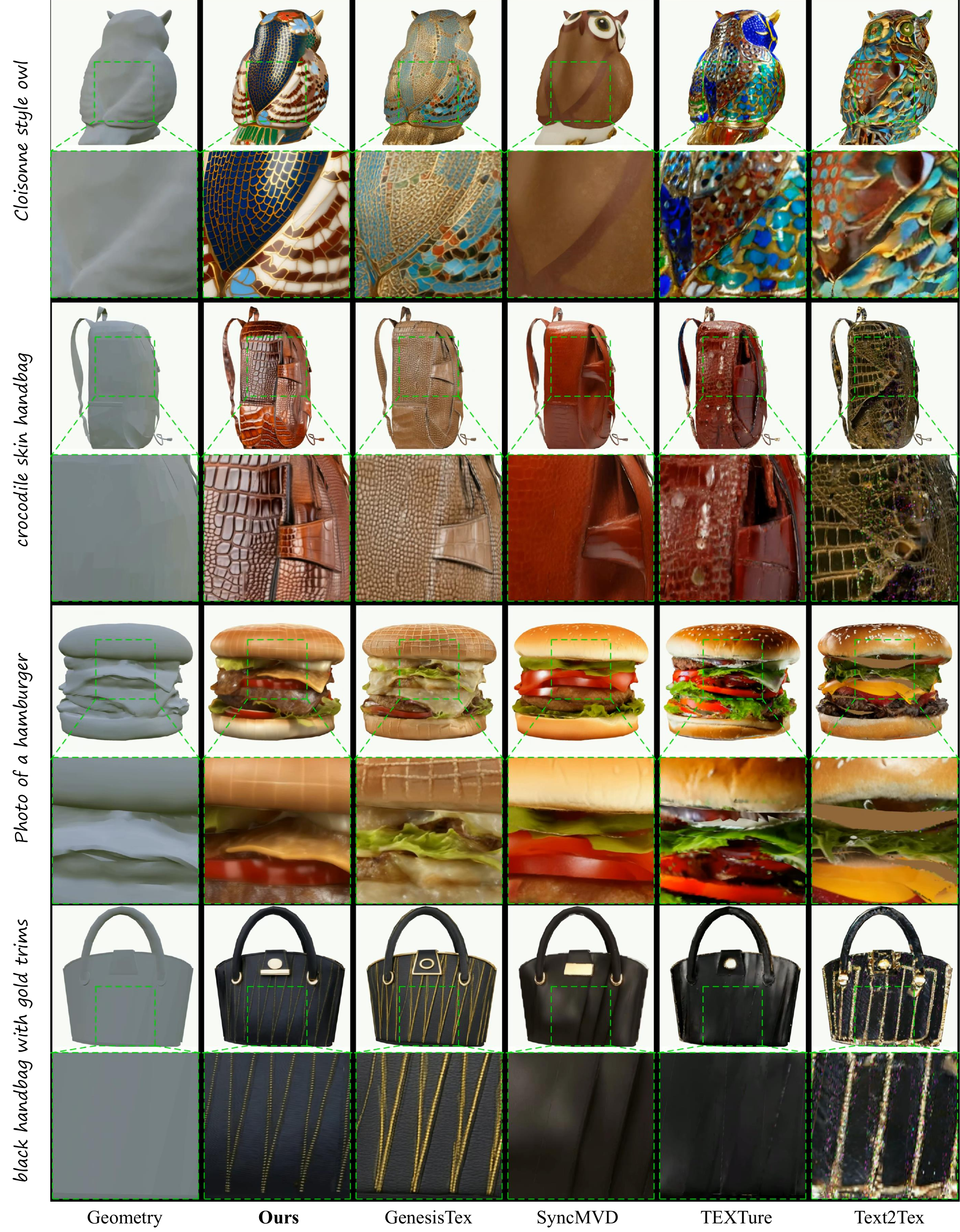}
  \caption{
    More comparison results with different methods. 
  }
  \label{fig:comparison3}
\end{figure*}

\begin{figure*}
  \centering
  \includegraphics[width=0.9\linewidth]{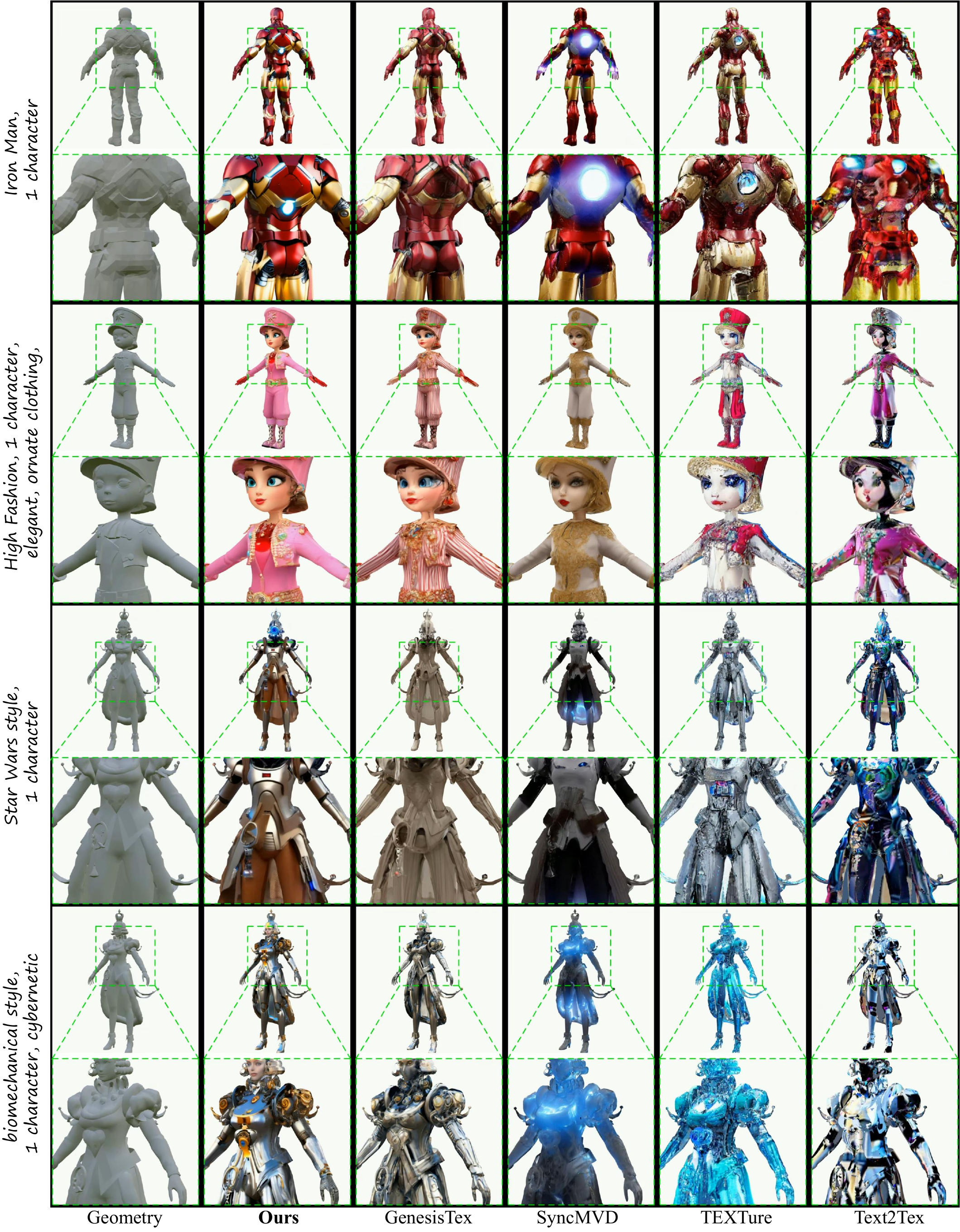}
  \caption{
    More comparison results with different methods.
  }
  \label{fig:comparison4}
\end{figure*}

\begin{figure*}
  \centering
  \includegraphics[width=0.9\linewidth]{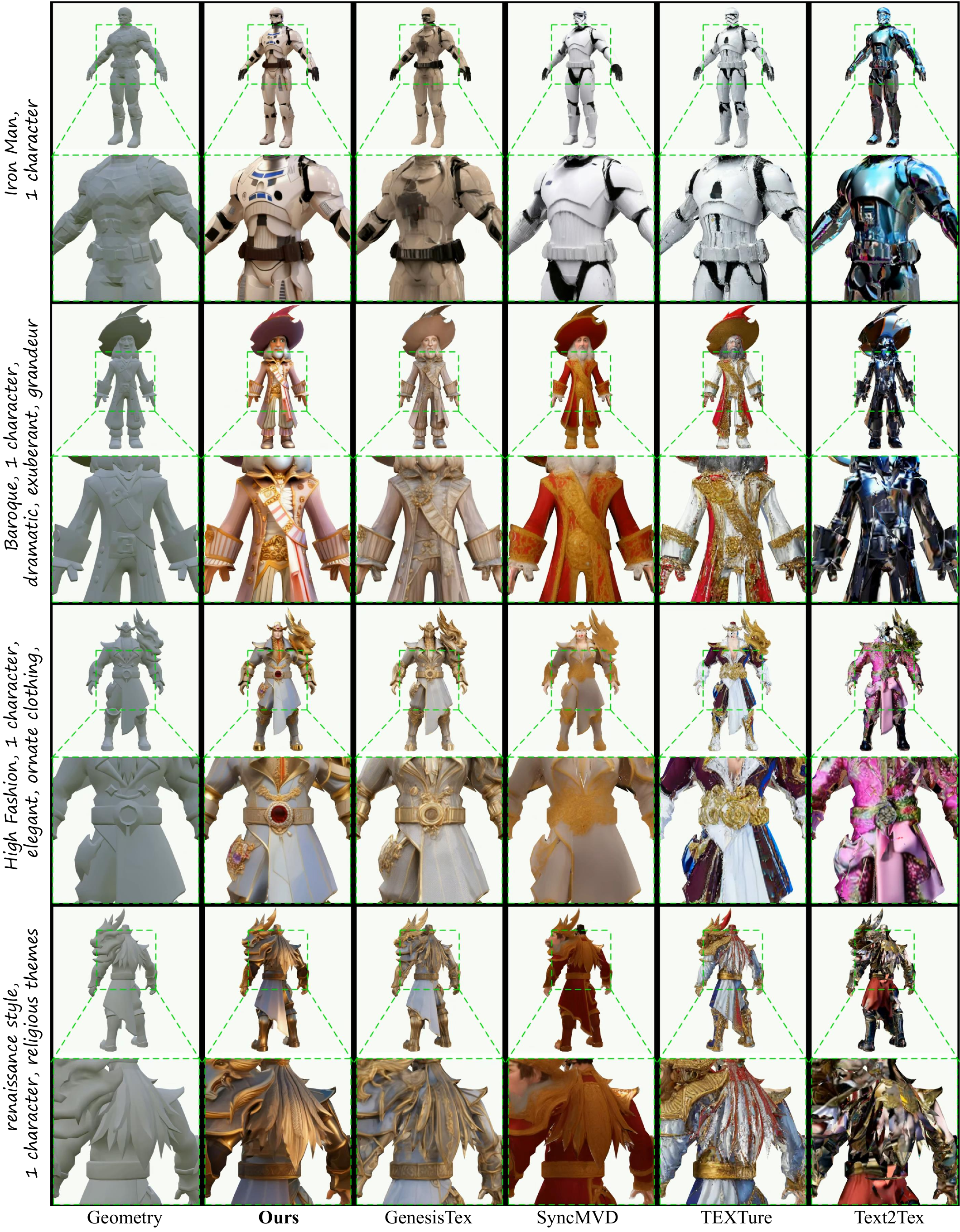}
  \caption{
    More comparison results with different methods.
  }
  \label{fig:comparison5}
\end{figure*}

We show extra ablation experiments on local attention in Fig.~\ref{fig:ablation_local_attn_2}. This example shows the ablation results on different attention mechanisms in multi-view generation without latent merge. Our local attention achieves the best multi-view consistency while preserving rich details close to the original unconstrained diffusion image(row 1).
In addition, we provide more comparison results with different methods in Fig.~\ref{fig:comparison3},~\ref{fig:comparison4},~\ref{fig:comparison5}. More results by our methods on various meshes and styles can be viewed in Fig.~\ref{fig:more_obj1},~\ref{fig:more_obj2},~\ref{fig:more1},~\ref{fig:more2},~\ref{fig:more3}. We highly recommend readers to watch the videos for more results.

\bibliography{aaai25}

\begin{figure}
  \centering
  \includegraphics[width=\linewidth]{figs/ablation/ablation_attn_2.pdf}
  \caption{
    Ablation results on different attention mechanisms in multi-view generation. Each view attends to its neighbors (top), each view attends to all other views (middle), our \textbf{local attention} (bottom) achieves the best multi-view consistency while preserving rich details.
  }
  \label{fig:ablation_local_attn_2}
\end{figure}

\begin{figure*}
  \centering
  \includegraphics[width=\linewidth]{figs/teaser/teaser4.pdf}
  \caption{
    More results on meshes from objaverse\cite{deitke2023objaverse}. 
  }
  \label{fig:more_obj1}
\end{figure*}

\begin{figure*}
  \centering
  \includegraphics[width=\linewidth]{figs/teaser/teaser5.pdf}
  \caption{
    More results on meshes from objaverse\cite{deitke2023objaverse}. 
  }
  \label{fig:more_obj2}
\end{figure*}

\begin{figure*}
  \centering
  \includegraphics[width=\linewidth]{figs/teaser/teaser1.pdf}
  \caption{
    More results on meshes from industrial games. 
  }
  \label{fig:more1}
\end{figure*}

\begin{figure*}
  \centering
  \includegraphics[width=\linewidth]{figs/teaser/teaser2.pdf}
  \caption{
    More results on meshes from industrial games.
  }
  \label{fig:more2}
\end{figure*}

\begin{figure*}
  \centering
  \includegraphics[width=\linewidth]{figs/teaser/teaser3.pdf}
  \caption{
    More results on meshes from industrial games.
  }
  \label{fig:more3}
\end{figure*}